\documentclass[10pt,twocolumn,letterpaper]{article}

\usepackage{iccv}              

\newcommand{\red}[1]{{\color{red}#1}}

\definecolor{iccvblue}{rgb}{0.21,0.49,0.74}
\usepackage[pagebackref,breaklinks,colorlinks,allcolors=iccvblue]{hyperref}

\def\paperID{15181} 
\def\confName{ICCV}
\def\confYear{2025}

\title{Disentangling 3D from Large Vision-Language Models\\for Controlled Portrait Generation}

\author{Nick Yiwen Huang$^*$\\
Brown University
\and
Akin Caliskan\\
Flawless AI
\and
Berkay Kicanaoglu\\
Flawless AI
\and
James Tompkin\\
Brown University
\and
Hyeongwoo Kim\\
Imperial College London
}

\usepackage{lipsum}

\usepackage{booktabs} 

\usepackage{colortbl} 
\usepackage{esvect}

\usepackage{wrapfig}

\usepackage[export]{adjustbox}

\usepackage{makecell}
\usepackage{multirow}

\usepackage{xcolor}
\usepackage{balance}
\usepackage{float}

\usepackage{amsmath}

\usepackage{microtype}

\usepackage{color}

\newcommand{\sysname}{CLIPortrait\xspace}

\definecolor{good}{rgb}{0.7725,0.8784,0.7059} 
\definecolor{maybe}{rgb}{0.984,0.898,0.839}
\definecolor{bad}{rgb}{0.9,0.75,0.75}
\definecolor{na}{rgb}{0.8,0.8,0.8}

\definecolor{LightGray}{rgb}{0.88,0.88,0.88}

\definecolor{LightCyan}{rgb}{0.88,1,1}
\definecolor{babyblue}{rgb}{0.54, 0.81, 0.94}
\definecolor{beige}{rgb}{0.96, 0.96, 0.86}
\definecolor{brass}{rgb}{0.71, 0.65, 0.26}

\newcommand{\noisevec}{\mathbf{z}}

\newcommand{\controlvec}{\mathbf{c}}
\newcommand{\CLIPvec}{\mathbf{c}}

\newcommand{\ray}{\vec{r}}
\newcommand{\density}{\sigma}
\newcommand{\volfeat}{\mathbf{f}}

\newcommand{\direction}{\vec{\omega}}

\newcommand{\tnear}{{t_\text{n}}}
\newcommand{\tfar}{{t_\text{f}}}

\usepackage[hang,flushmargin]{footmisc}

\newcommand\blfootnote[1]{%
  \begingroup
  \renewcommand\thefootnote{}\footnote{#1}%
  \addtocounter{footnote}{-1}%
  \endgroup
}

\begin{document}
\maketitle
\blfootnote{*: The work was completed while Nick Huang was an intern at Flawless AI and back at Brown.}
\begin{abstract}

We consider the problem of disentangling 3D from large vision-language models, which we show on generative 3D portraits. This allows free-form text control of appearance attributes like age, hair style, and glasses, and 3D geometry control of face expression and camera pose. In this setting, we assume we use a pre-trained large vision-language model (LVLM; CLIP \cite{radford2021learning}) to generate from a smaller 2D dataset with no additional paired labels and with a pre-defined 3D morphable model (FLAME \cite{FLAME:SiggraphAsia2017}).
First, we disentangle using canonicalization to a 2D reference frame from a deformable neural 3D tri-plane representation. But, another form of entanglement arises from the significant noise in the LVLM's embedding space that describes irrelevant features. This damages output quality and diversity, but we overcome this with a Jacobian regularization that can be computed efficiently with a stochastic approximator. 
Compared to existing methods, our approach produces portraits with added text and 3D control, where portraits remain consistent when either control is changed.
Broadly, this approach lets creators control 3D generators on their own 2D face data without needing resources to label large data or train large models.
\end{abstract}

\vspace{-15pt}
\section{Introduction}
\label{sec:intro}


Research into portrait generation now lets us create realistic 3D images via machine learning from photograph data-sets, with use in visual effects, games, and virtual reality. However, the problem of how to control the generation process to meet desired face attributes remains open. Such attributes may span hair color, face shape or expression, or age or hair style. Ideally, all of these attributes would be controllable independently so that, say, editing the hair style of a person does not change their expression.

\begin{figure}[t]
  \centering
  \small
  \setlength{\tabcolsep}{0pt}
  \renewcommand{\arraystretch}{0.0}
  \scalebox{.65}{
  \begin{tabular}{ccc|ccc}

  \multicolumn{3}{c}{``This man has bags under eyes''} & \multicolumn{3}{c}{``This man has bags under eyes. He has beard''} \\ 
  {\includegraphics[width=0.25\linewidth]{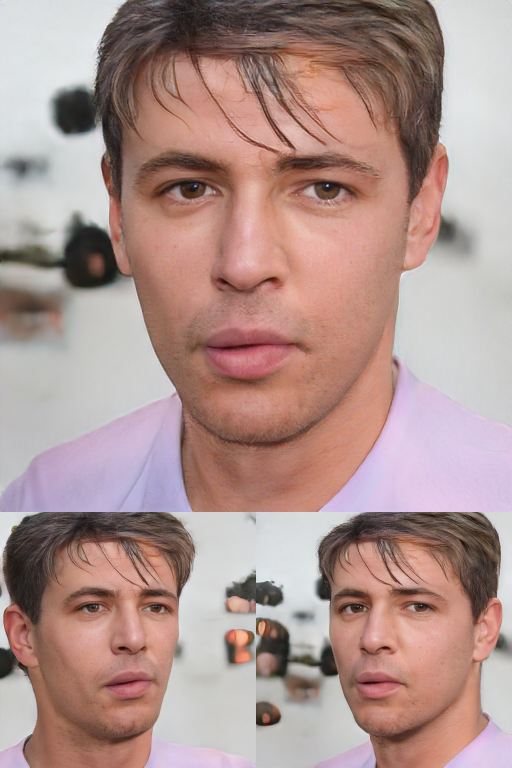}} & 
  {\includegraphics[width=0.25\linewidth]{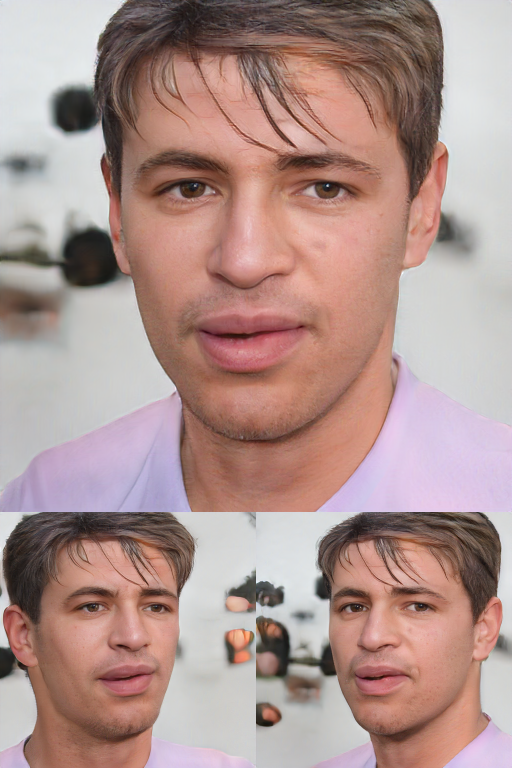}} &
  {\includegraphics[width=0.25\linewidth]{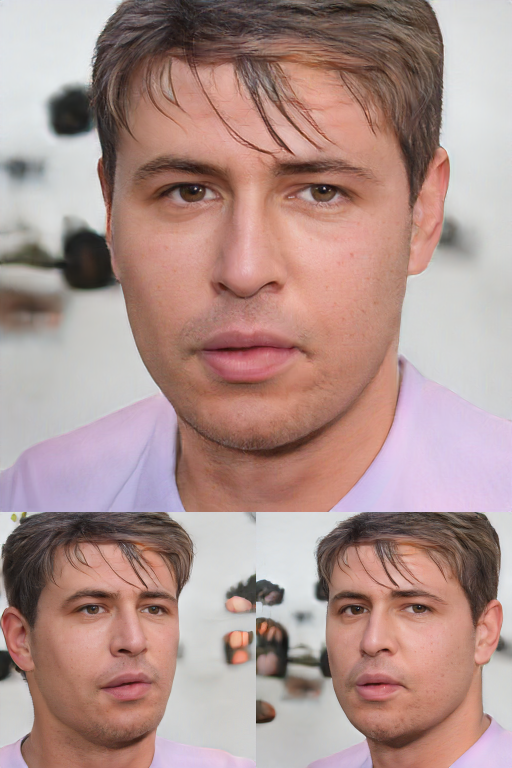}}
  & 
  {\includegraphics[width=0.25\linewidth]{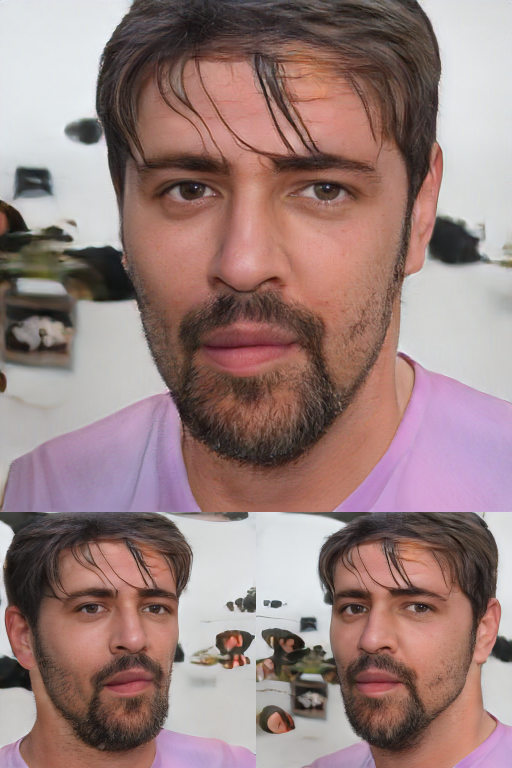}} & 
  {\includegraphics[width=0.25\linewidth]{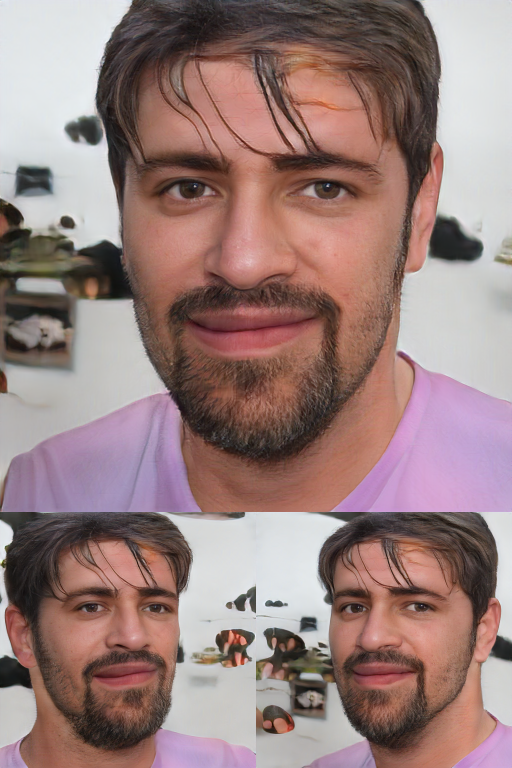}} &
   {\includegraphics[width=0.25\linewidth]{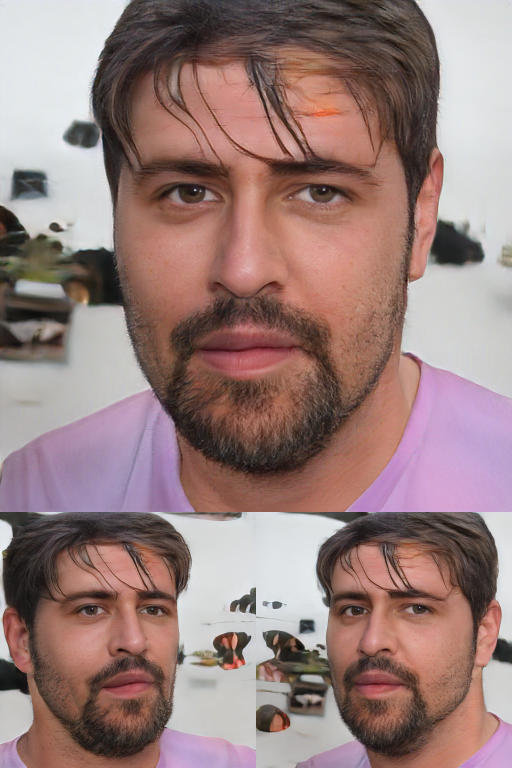}}
  \\
  \midrule
    3D Face  & \multicolumn{1}{c}{Expression} & Shape & 3D Face  &  \multicolumn{1}{c}{Expression} & Shape \\
    \midrule
    Generation & \multicolumn{2}{c}{Manipulations} & Generation & \multicolumn{2}{c}{Manipulations}
  \end{tabular}
  }
  \caption{\sysname allows text-guided 3D portrait generation and editing of 3D portraits. Given input text, \sysname can synthesize high-quality 3D faces with disentangled geometry and camera control using parametric 3D face models.}
  \label{fig:teaser}
  \vspace{-0.5cm}
\end{figure}

Such controls are typically induced during learning through labels. Existing methods have focused on two label modes: 3D morphable models (3DMM) and text. 3DMMs are linear statistical models obtained from precisely-aligned 3D scan datasets~\cite{bfm09,FLAME:SiggraphAsia2017}. Fitting a 3DMM to a dataset allows conditioning a generator for precise control~\cite{bergman2022gnarf}, but most 3DMMs are of the face or head only and so offer no appearance variation/control outside the 3DMM's domain. Text is less precise but may allow easier high-level control, including all of the appearance. One approach is to pair matching photos and text labels of attributes, e.g., hair color, eye color, wearing glasses. However, every new attribute requires labeling, leading to a limited set of controls, limited sample size, or limited sample diversity. For instance, each photo in FFHQ-Text~\cite{zhou2021generative} has 9 text annotations, each describing a subset of 162 attributes for detailed geometry and appearance. But, the dataset only covers women, and has only 760 photos in total.

An alternative approach is to train a large vision-language model (LVLM) on many millions of in-the-wild photos and captions to define latent spaces that correlate text and images, e.g., the popular CLIP~\cite{radford2021learning} model's public release makes accessible a model that would be too costly to train for many. 
But, these models can be inconsistent: as each photo is not labeled with all desired text attributes, different geometry and appearance attributes end up \emph{entangled} due to spurious correlations. Training a generator using CLIP is a challenge: changing one attribute invariably changes another---an unsatisfying interaction.

To alleviate these limitations, we propose a method generate 3D portraits using both CLIP-derived text and 3D conditioning. This produces comparable quality and diversity of output to unconditional models but still allows independent control of geometry and appearance attributes across low, mid, and high levels. Our model is trained on a database of \emph{unlabeled} 2D face photos (e.g., FFHQ~\cite{karras2019style}), using the pretrained LVLM CLIP and the 3DMM FLAME. Such an approach requires disentangling CLIP itself to isolate control over all parameters that could be controlled by FLAME, without damaging the ability of CLIP to describe diverse portraits---na\"{i}ve approaches lead to low image quality, low sample diversity, or limited control.
To disentangle, we learn to deform database faces to both 1) a 3D canonical space represented by a neural tri-plane, and to 2) a 2D canonical space in which CLIP can provide more reliable pseudo-labels that align text to images. In this way, CLIP only has to describe the appearance unexplained by the deformation and projection of a 3D volume into a camera. To bypass per-sample optimization, we define lightweight attribute mixing functions that can be baked from CLIP text prompts, e.g., `blue eyes', `blonde hair', for fast editing. 

Beyond providing a model for high-quality controllable generation of 3D portraits, our approach more broadly defines a method to allow creators without the compute resources to train an LVLM directly to instead adapt one to their own smaller 2D face data, such as proprietary data from games or VFX studios, to allow text and fine-grained geometry control without expensive labeling.

It is worth noting that our contributions are orthogonal to specific LVLMs and 3D generative models. While we pick CLIP~\cite{radford2021learning} as our LVLM and GNARF~\cite{bergman2022gnarf} as the generative backbone given their readily availability, our contributions are still applicable if,~\eg, we swap CLIP with LLaVA~\cite{liu2024visual} or replace GNARF with gaussian splatting models~\cite{kirschstein2024gghead,chu2024generalizable,hyun2024adversarial} or 3D diffusion models~\cite{zhou2024diffgs,lan2023gaussian3diff} as long as they allow deformation for explicit geometry control.

\section{Why LVLMs Struggle as 3D Labelers}
\label{sec:background}
Given a set of unlabeled images $\hat{x}\in\mathcal{X}$, we aim to construct a generator $G(\controlvec,\noisevec):\mathcal{C}\times\mathcal{Z}\rightarrow\mathcal{X}$ where $\controlvec\in\mathcal{C}$ denotes factors that allow us to control the generation process, and $\noisevec\in\mathcal{Z}$ denotes a noise vector accounting for the unexplained factors of variation in the synthesized appearance. 
To simply discussion, let $\controlvec=\left[\controlvec_\text{cam}, \controlvec_\text{geo}, \controlvec_\text{imp}\right]$ where the camera pose $\controlvec_\text{cam}$ and the geometry $\controlvec_\text{geo}$ are explicitly controllable in a deformable 3D generative model. Anything we want to implicitly control by text, we leave in $\controlvec_\text{imp}$.
The goal is to induce $\controlvec_\text{imp}$ from free-form text prompts $t\in\mathcal{T}$ to create photorealistic face images $x$ that aligns with $\controlvec_\text{imp}$ while preventing $\controlvec_\text{imp}$ from a) interfering $\controlvec_\text{cam}$ and $\controlvec_\text{geo}$. b) overshadowing $z$.
Our contributions focus on disentanglement; to explain this, first we consider why entanglement arises in LVLMs via their \emph{alignment} objective.

\vspace{-10pt}
\paragraph{Alignment} Given text $t$, the alignment objective requires that the generated sample $x$ be semantically consistent with $t$. Let $E_\text{txt}(t):\mathcal{T}\rightarrow\mathcal{R}$ be an encoder that maps text $t$ to some representation $\textbf{r} \in \mathcal{R}$; 
likewise, $E_\text{img}(x):\mathcal{X}\rightarrow\mathcal{R}$ maps an image $x$ to the same space $\mathcal{R}$. We can define the alignment of $x$ to $t$ as maximizing the mutual information $I(x;t)$, which is bounded by the mutual information $I(\textbf{r}_x;\textbf{r}_t)$ between $\textbf{r}_x=E_\text{img}(x)$ and $\textbf{r}_t=E_\text{txt}(t)$. 

For CLIP, $E_\text{img}$ and $E_\text{txt}$ are trained with the InfoNCE objective. Oord et al.~\cite{oord2018representation} show that minimizing InfoNCE maximizes the lower bound of $I(\textbf{r}_x;\textbf{r}_t)$, given text-image pairs $(x,t)\in\mathcal{Y}$:%
\vspace{-0.4cm}
\begin{small}
\begin{align}
\text{InfoNCE}(x,t) &=-\mathbb{E}_{(x,t)\sim\mathcal{Y}}\left[\log \frac{\exp(\cos(\textbf{r}_x,\textbf{r}_t))}{\sum_{\substack{x'\sim\mathcal{X}\\t'\sim\mathcal{T}}} \exp(\cos( \textbf{r}_{x'}, \textbf{r}_{t'} )) } \right] \\
&\geqslant -I(\textbf{r}_x;\textbf{r}_t)+\text{constant}
\end{align}
\end{small}

\noindent CLIP was trained on Internet-scale 2D image/text data $\mathcal{Y}$. For a target dataset $\mathcal{X}$ for which we would like a generator, say, the high-quality close-ups in FFHQ, let's assume that CLIP happens to cover all portrait images. Then, text-guided 2D generation becomes viable even though FFHQ has no text labels: we can use $\controlvec_\text{imp}=\textbf{r}_x=E_\text{img}(x_{\text{FFHQ}})$ to condition $G$ during training, which requires no text labels. Then, at inference time, we replace $\textbf{r}_x$ by $\textbf{r}_t$---this allows $G$ to generate an image from any text prompt provided by the user. The contrastive objective while brings $\textbf{r}_x$ and $\textbf{r}_t$ as close as possible, does not fully eliminate the modality gap in $\mathcal{R}$~\cite{liang2022mind} and instead causes entanglement in $\textbf{r}_x$ for our concerns.

\vspace{-10pt}
\paragraph{Entanglement.} $\mathcal{Y}$ encompasses images of all things on the Internet---not just portraits with detailed text captions. This situation has two problems. 

1) Only a small fraction of portraits in $\mathcal{Y}$ contain a description of geometry, and the description is coarse, e.g., \textit{`smile'} does not define how wide the smile is, or \textit{`viewed from side on'} does not define the 3D camera angle. This allows $E_\text{img}$ to encode incomplete geometry information $\textbf{r}_{x_\text{geo}}$ and camera information $\textbf{r}_{x_\text{cam}}$ in $\textbf{r}_x$ as the result of spurious correlations,~\eg celebrity faces are more likely to have a front camera pose and smiley expression. As such, introducing a 3D representation within the generator and then using $\textbf{r}_x$ as $\controlvec_\text{imp}$ leads to poor results since $\textbf{r}_{x_\text{geo}}$ and $\textbf{r}_{x_\text{cam}}$ are strongly at odds with $\CLIPvec_\text{cam}$ and $\CLIPvec_\text{geo}$. For instance, if $\CLIPvec_\text{cam}$ specifies a camera pose different than what $\textbf{r}_{x_\text{cam}}$ dictates, $G$ is ill-behaved as it receives conflicting conditions for the same control. 
We address this conflict by proposing 2D and 3D canonicalizations to eliminate $\textbf{r}_{x_\text{cam}}$ and $\textbf{r}_{x_\text{geo}}$ from $\textbf{r}_x$. As long as all instances of $x$ share the same camera pose and geometry, the generation process can no longer distinguish such information from $\textbf{r}_x$.

2) The contrastive objective forces $\textbf{r}_x$ to be as discriminative as possible in $\mathcal{Y}$, but the most discriminative factors in $\textbf{r}_x$ could be irrelevant to portraits---non-portrait noise factors $\textbf{r}_{x_\text{noise}}$ can outweigh any useful factors when using $\textbf{r}_x$ as $\controlvec_\text{imp}$. Examples of $\textbf{r}_{x_\text{noise}}$ include the camera used to produce the image, image format and quality, possible geographical location where the picture was taken (Fig.~\ref{fig:clipnoise}). We show in the supplemental that these noise factors can considerably overshadow factors that describe the person in the image. Since $\textbf{r}_{x_\text{noise}}$ is typically the most discriminative factors, it tends to be unique and vary greatly for different instances of $x$. As a result, this encourages $G$ to be a deterministic function of $\textbf{r}_x$ and completely ignore $z$, since each $x$ can be uniquely identified by just $\textbf{r}_{x_\text{noise}}$ (and therefore $\textbf{r}_x$). We avoid this degeneracy by introducing a Jacobian regularization that penalizes the generator's sensitivity to $\textbf{r}_{x_\text{noise}}$.

\begin{figure}[t]
    \centering
    \includegraphics[trim={12.85cm 1cm 0 0},clip,width=.5\textwidth]{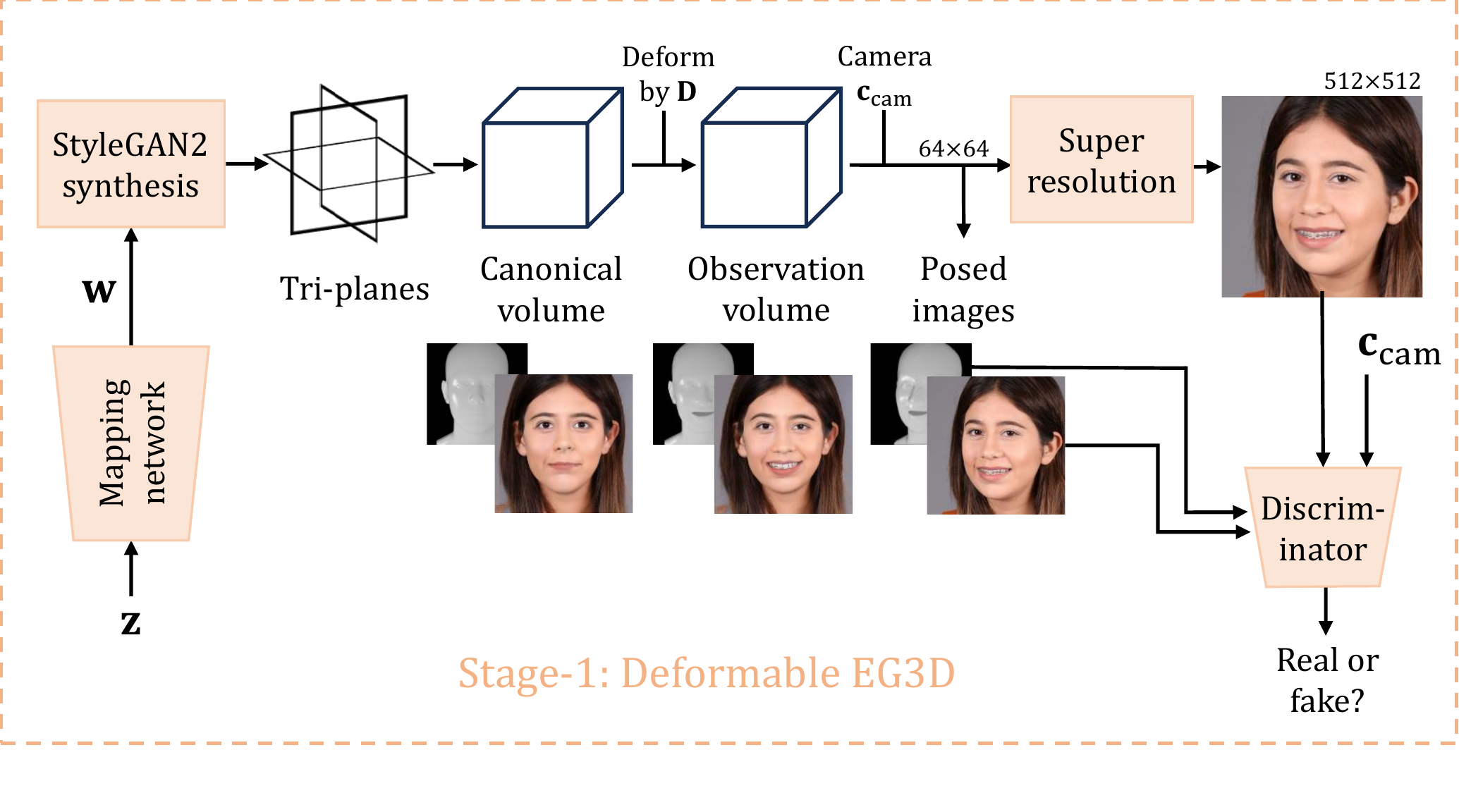}
    \vspace{-0.75cm}
    \caption
    {\textbf{Overview of Stage-1.} We use a deformable 3D generative model for Stage-1. We adapt GNARF~\cite{bergman2022gnarf} to improve the geometry conditioning of the discriminator using mesh renders.}
    \vspace{-15px}
    \label{fig:overview_stage_1}
\end{figure}%

\vspace{-8pt}
\section{Method}
\label{sec:method}

\subsection{High-level Overview}

\begin{figure*}[t]
    \centering
    \includegraphics[clip,trim={0cm 1.9cm 0cm 0cm},width=.8\textwidth]{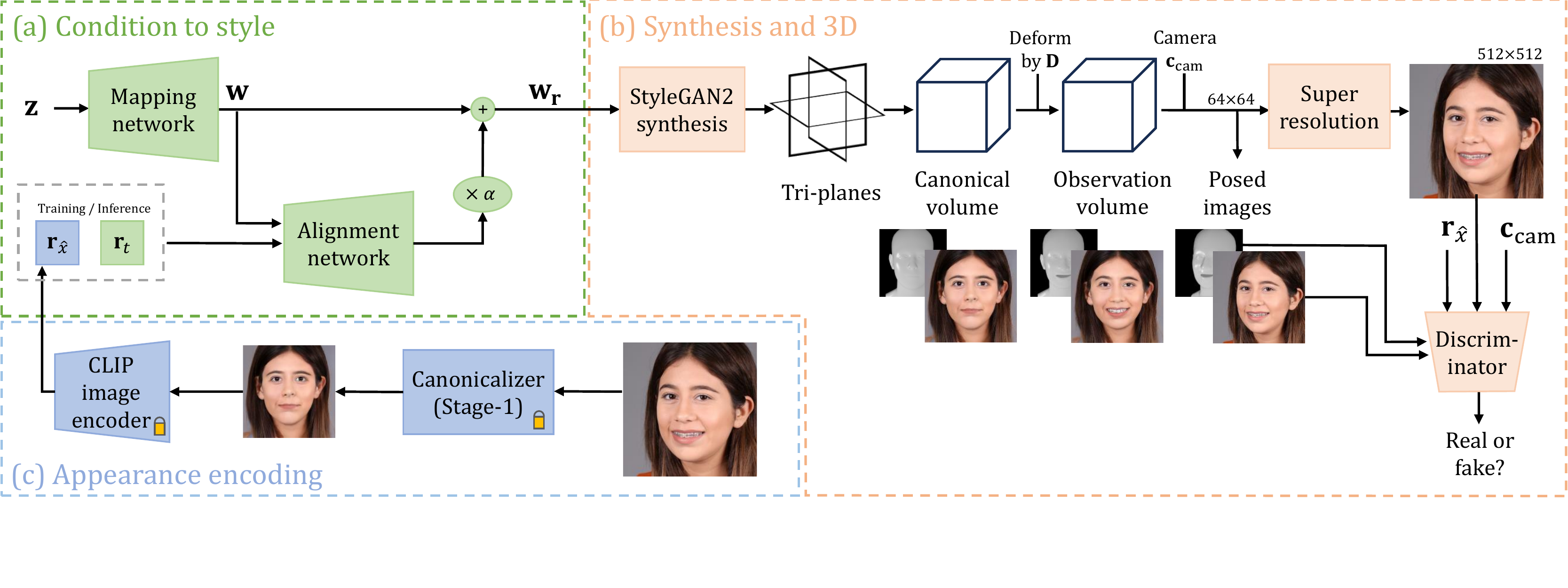}
    \caption
    {\textbf{Overview of Stage-2.} \textbf{(a)} The conditioning networks remap CLIP embeddings to a space $w_\textbf{r}$ in which 3D information is ignored to be considered in stage (b), while noise vector $\noisevec$ maintains sample diversity. \textbf{(b)} We synthesize a neural radiance volume via a tri-plane in a 3D canonical space, and with a particular appearance defined by $w_\textbf{r}$. Then, we deform this volume by FLAME parameters after fitting to the dataset. A discriminator judges the rendered output image. \textbf{(c)} The alignment network in (a) can only achieve a $w_\textbf{r}$ free of 3D information if all generated images are `3D equivalent'; we achieve this via canonicalization.}
    \vspace{-10pt}
    \label{fig:overview}
\end{figure*}%

Our overall approach is a 3D GAN (Fig.~\ref{fig:overview}) that uses two training stages to disentangle CLIP for an unlabeled observation dataset $\mathcal{X}$. These two stages are necessary because our full model requires an unconditional deformable generator to obtain the canonicalized appearance condition $\textbf{r}_{\hat{x}}$. Furthermore, bootstrapping our full model from an unconditional model helps avoid degenerate solutions caused by $\textbf{r}_{x_\text{noise}}$ which we show in Section~\ref{sec:regularization}.

\noindent \textbf{Stage-1 (Unconditional Generation)} 
First, we train a 3D generator $G$ with no \textit{a priori} text understanding to output a volumetric tri-plane representation (Fig. \ref{fig:overview_stage_1}).
This representation is deformed from its canonical format using a 3D map \textbf{D} derived from 3DMM, and projected back to an intermediate low-resolution image $x^{\ddag}$ according to camera parameters $\controlvec_\text{cam}$ (Sec.~\ref{sec:deformable_gen}).
Finally, we use a super-resolution module to produce a sample $x$ at desired target resolution from $x^{\ddag}$.
To train Stage-1, we use a discriminator to assess whether the rendered image $x$ is real or fake and use an adversarial objective to optimize model $G$.

\noindent \textbf{Canonicalization} 
Once $G$ is trained, we can canonicalize each sample $x$ to a fixed-geometry 3D volume by inverting $\textbf{D}$, and then render a frontal 2D image $\hat{x}$ by inverting $\controlvec_\text{cam}$. We run pre-trained CLIP on each canonicalized 2D image $E_\text{img}(\hat{x})$ to compute $\textbf{r}_{\hat{x}}$. 

\noindent \textbf{Stage-2 (Conditioning Appearance on Text)} To add text control, we use an alignment network $T_G$ to modify any random style vector $w$ according to $\textbf{r}_{\hat{x}}$. We disentangle 3D information from CLIP using 2D canonicalization such that $\textbf{r}_{\hat{x}}$ only contains frontal appearance information. The alignment network preserves the randomness of $w$, which maintains output diversity and enables interactive style mixing. Disentanglement of 3D information from CLIP occurs because $\textbf{r}_{\hat{x}}$ is predicted from 2D images that all share the same camera and geometry; any possible means to distinguish such information has been factored out of the generator.



\subsection{Generating Deformable 3D Portraits}
\label{sec:deformable_gen}
We use a tri-plane reduction of a neural radiance field, like EG3D~\cite{Chan2022}. Given a 3D point, we query the tri-plane for a feature vector $\textbf{f}^\prime$, and obtain volumetric features $\volfeat$ and density $\density$ from $\textbf{f}^\prime$ using an MLP. We obtain a pixel of the low resolution render by integrating $\volfeat$ and $\density$ along ray $\ray$:
\begin{small}
\begin{align}
F(\ray)=\int_{\tnear}^{\tfar}T(t)\density(t)\volfeat(t)\text{d}t, \hspace{0.1cm} \text{where} \hspace{0.1cm} 
T(t)=\exp\left( -\int_{t_n}^{t_f}\density(s)\text{d}s \right)
\label{eq:volumerendering}
\end{align}
\end{small}
where $\tnear$ and $\tfar$ are the near and far bounds of the ray $\ray(t)=\textbf{o}+\direction t$ along the direction $\direction$ from the origin $\textbf{o}$. 
%
%
Precise camera control is possible by changing the rays $\ray$ along which $F$ is aggregated. To control portrait geometry, including face shape and facial expression, we deform the ray along which $F$ is aggregated:
\begin{align}
\volfeat(\textbf{x}')=\volfeat(\textbf{D}(\textbf{x})),
\label{eq:deformvolume}
\end{align}
where $\textbf{x}$ is a coordinate in the observation space and $\textbf{D}:\mathbb{R}^3\rightarrow\mathbb{R}^3$ is a deformation that maps $\textbf{x}$ to a canonical space. Deforming coordinates from the canonical space removes the need for generator $G$ to represent varying geometry.

$\textbf{D}$ can be constructed from explicit 3DMMs. We use FLAME ~\cite{FLAME:SiggraphAsia2017}: it has controllable parameters $\mathbf{\beta}\in\mathbb{R}^{100}$ for face shape, $\mathbf{\theta}\in\mathbb{R}^{6}$ for jaw and head pose, and $\mathbf{\psi}\in\mathbb{R}^{50}$ for facial expression. We estimate these for the observation mesh from $\hat{x}$ using DECA~\cite{DECA:Siggraph2021}. We construct $\textbf{D}$ from FLAME analytically using the surface field (SF) method in GNARF by Bergman et al.~\cite{bergman2022gnarf}. SF derives the deformation field from the canonical mesh and the observation mesh aligned to $\hat{x}$. For the canonical space, we set the FLAME shape and expression coefficients to 0, but leave the jaw open to synthesize the mouth interior.

Given a target coordinate $\textbf{x}$, SF locates its nearest triangle $t_\textbf{x}^D=[\textbf{v}_0,\textbf{v}_1,\textbf{v}_2]\in\mathbb{R}^{3\times3}$ on the target mesh, and computes the barycentric coordinates $[u,v,w]$ of the projection of $\textbf{x}$ on $t_\textbf{x}^D$. To calculate the deformed coordinate, we retrieve the corresponding triangle $t_\textbf{x}^C$ on the canonical mesh and its normal $\textbf{n}_{t_\textbf{x}}^C$:
\begin{small}
\begin{align}
\textbf{D}(\textbf{x})=t_\textbf{x}^C\cdot[u,v,w]^\top+\left \langle \textbf{x}-t_\textbf{x}^D\cdot[u,v,w]^\top,\textbf{n}_{t_\text{x}}^D \right \rangle\textbf{n}_{t_\text{x}}^C
\end{align}
\end{small}
which we use to query the canonical volume. Since geometry variations are explicitly controlled by deformation and have been factored out from $G$, entanglement between facial expression and camera pose in EG3D no longer exists; as such, we do not use generator pose conditioning~\cite{Chan2022}.


Without informing the discriminator of the deformation, there is no guarantee that the deformed volume matches the expected deformation. To condition the discriminator, Bergman et al.~\cite{bergman2022gnarf} concatenate the camera pose with FLAME parameters. However, this leads to training instability, and sample quality is severely degraded even with the noise perturbation trick~\cite{bergman2022gnarf}. As Huang et al.~showed~\cite{Huang_2024_WACV}, this is because the conditioning vector depends upon the unknown PCA basis for FLAME, making it difficult for the optimization to use this additional input. 
Instead, we adopt the Huang et al.~method. We texture the mesh with its vertex coordinates in world space. Then, we render the observation mesh under $\controlvec_\text{cam}$ and concatenate the render $rdr$ with $x$ as input to the discriminator. We observe no training instability or quality degradation using this conditioning.

\subsection{Canonicalization}
Given our deformable EG3D, canonicalization can be reduced to an inversion problem. Specifically, the sample generation process of the deformable generator is given by:
\begin{small}
\begin{align}
w=M_G(\noisevec)\\
f=S_G(w)\\
x=V(f,\controlvec_\text{cam},\controlvec_\text{geo})
\end{align}
\end{small}
where $M_G$ and $S_G$ are the style mapping and synthesis networks of $G$, $V$ denotes deformable volume rendering. For each image $x$ in the training set, we estimate $\controlvec_\text{cam}$ and $\controlvec_\text{geo}$ using off-the-shelf models. The corresponding latent vector $w_{x}$ can then be obtained by solving the following optimization problem:
\begin{align}
w_{x} = \underset{w}{\mathrm{argmin}}\,\text{D}_\text{LPIPS}\left(V(S_G(w),\controlvec_\text{cam},\controlvec_\text{geo}),x\right )
\end{align}
where $\text{D}_\text{LPIPS}$ denotes the LPIPS distance that we use as our image similarity. Given $w_{x}$, we can now re-render the canonicalized $\hat{x}$ under neutral camera pose and neutral geometry. We define the neutral camera pose $\controlvec_\text{n\_cam}$ to be fully frontal and the neutral geometry $\controlvec_\text{n\_geo}$ to have canonical FLAME parameters. The canonicalized $\hat{x}$ is given by:
\begin{small}
\begin{align}
\hat{x}=V(S_G(w_{x}),\controlvec_\text{n\_cam},\controlvec_\text{n\_geo})
\end{align}
\end{small}
and the disentangled condition vector $\controlvec_\text{imp}=\textbf{r}_{\hat{x}}=E_\text{img}(\hat{x})$. Note that canonicalization happens before Stage-2 training as a data preprocessing step, and so it has no impact on the training time of Stage-2.

\subsection{Conditioning on LVLM Text}
\begin{figure}[t]
    \centering
    \small
    \scalebox{0.8}{
    \setlength{\tabcolsep}{1pt}
    \renewcommand{\arraystretch}{0.4}
  \begin{tabular}{cccc|cccc}
    \multicolumn{4}{
c}{\bf Full Model} &  \multicolumn{4}{c}{\bf Without Canonicalization }   \\ 
    
    \includegraphics[valign=m,width=0.14\linewidth]{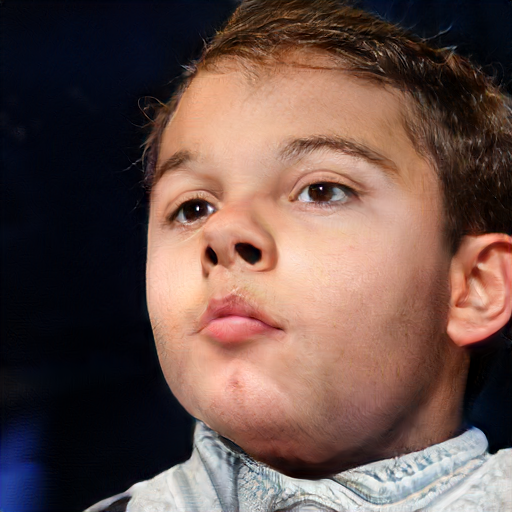} & 
    \includegraphics[valign=m,width=0.14\linewidth]{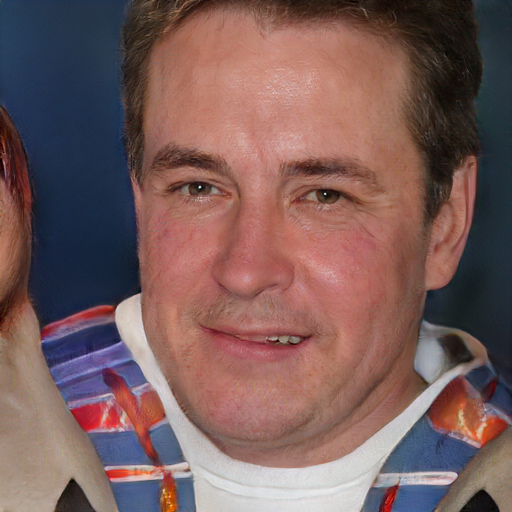} &  
    \includegraphics[valign=m,width=0.14\linewidth]{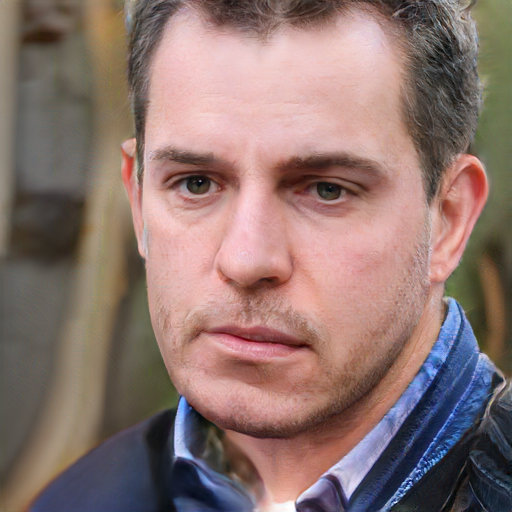} &
    \includegraphics[valign=m,width=0.14\linewidth]{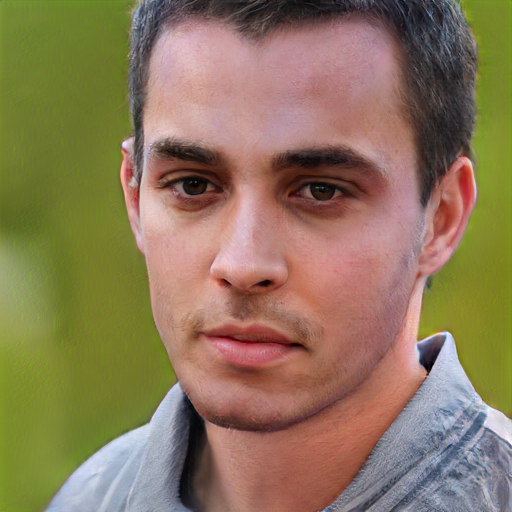} &
    \includegraphics[valign=m,width=0.14\linewidth]{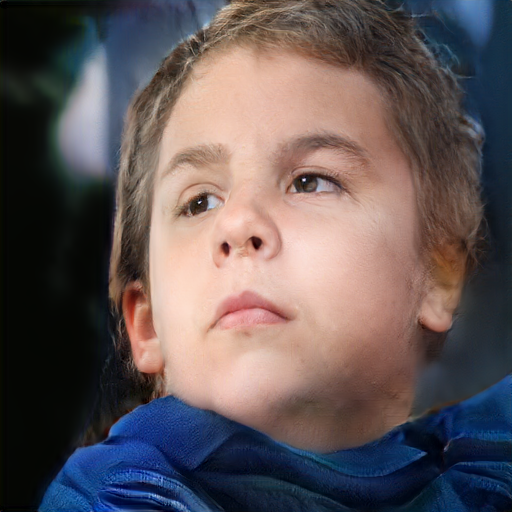} &
    \includegraphics[valign=m,width=0.14\linewidth]{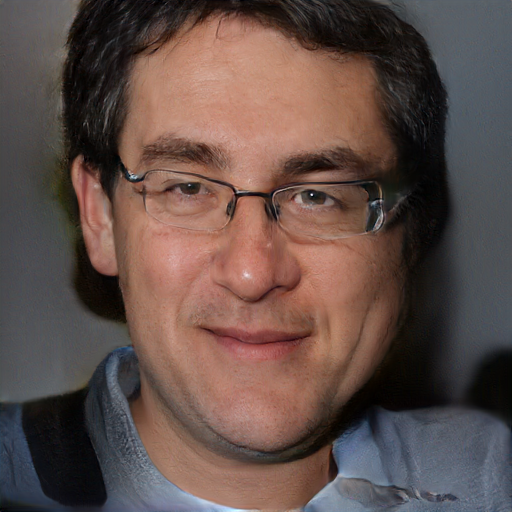} &
    \includegraphics[valign=m,width=0.14\linewidth]{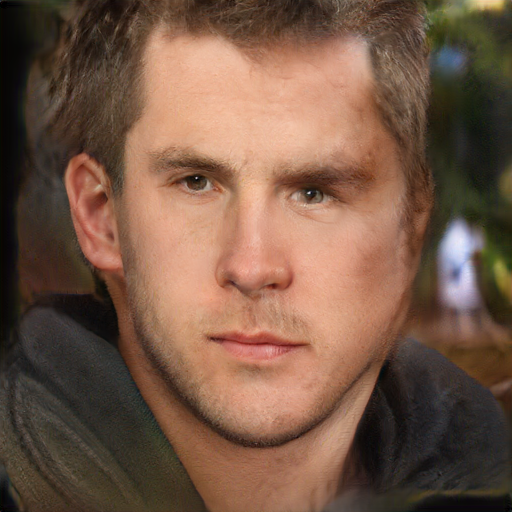} &
    \includegraphics[valign=m,width=0.14\linewidth]{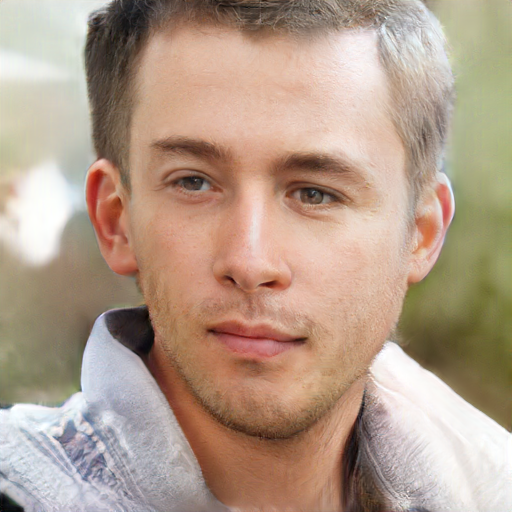}   \\
    
    
    \includegraphics[valign=m,width=0.14\linewidth]{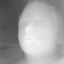} & 
    \includegraphics[valign=m,width=0.14\linewidth]{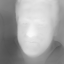} &  
    \includegraphics[valign=m,width=0.14\linewidth]{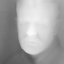} &
    \includegraphics[valign=m,width=0.14\linewidth]{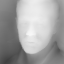} &
    \includegraphics[valign=m,width=0.14\linewidth]{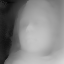} &
    \includegraphics[valign=m,width=0.14\linewidth]{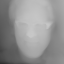} &
    \includegraphics[valign=m,width=0.14\linewidth]{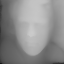} &
    \includegraphics[valign=m,width=0.14\linewidth]{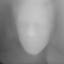}
    
  \end{tabular}
    }
    \caption
    {
    Without canonicalization before CLIP, faces look flat. Depth renderings of the estimated volume underneath show the more distorted space. [FFHQ 5mil.~images.] This directly shows our major insight: the CLIP embedding of the uncanonicalized image inherently includes camera pose information that must be disentangled, else it confuses the 3D generation.
    }    
    \label{fig:ablation_cano}
    \vspace{-5pt}
\end{figure}

We first visualize the importance of canonicalization in Figure \ref{fig:ablation_cano}: using the CLIP embedding from raw training images $x$, generation tries to incorrectly flatten faces to stay frontal regardless of the camera pose. Whereas conditioning on the canonicalized $\textbf{r}_{\hat{x}}$ produces correct geometry since neither $G$ nor $D$ can cheat with $\textbf{r}_{\hat{x}}$ for camera pose and geometry, and they must rely on $\controlvec_\text{cam}$ and $\controlvec_\text{geo}$ for such information.


To condition on $\textbf{r}_{\hat{x}}$, it is possible to either train a new model from scratch that takes $\textbf{r}_{\hat{x}}$ as an input, or adapt our unconditional deformable model from Stage 1 to handling $\textbf{r}_{\hat{x}}$. We choose the latter as we show in the following section that direct generation from $\textbf{r}_{\hat{x}}$ is prone to degenerate solutions due to the presence of $\textbf{r}_{\hat{x}_\text{noise}}$. Our Stage 2 model splits conditional generation into two easier steps: unconditional generation (\ie Stage 1) and alignment, where the latter seeks to modify an existing random sample such that it aligns with $\textbf{r}_{\hat{x}}$. Note that our Stage 1 model already facilitates well-behaved unconditional generation, as long as the alignment step is also well-behaved, we obtain well-behaved conditional generation on $\textbf{r}_{\hat{x}}$. 

Toward this goal, we introduce a CLIP alignment network $T_G$ to $G$ which predicts a personalized direction for the style vector $w$ of a random sample, along which the sample gains alignment toward $\textbf{r}_{\hat{x}}$:
\begin{small}
\begin{align}
w_{\textbf{r}_{\hat{x}}}(\alpha)=w+\alpha T_G(w,\textbf{r}_{\hat{x}})
\label{eq:alpha_weight}
\end{align}
\end{small}
$\alpha\in[0,1]$ is a scalar that controls the alignment strength, unless otherwise specified, $w_{\textbf{r}_{\hat{x}}}$ implies $\alpha=1$. Similarly, we introduce an alignment network $T_D$ to the discriminator $D$ as follows:
\begin{small}
\begin{align}
u=S_D(x,rdr)\\
v=M_D(\controlvec_\text{cam})\\
v_{\textbf{r}_{\hat{x}}}(\alpha)=v+\alpha T_D(v,\textbf{r}_{\hat{x}})\\
D(x\mid rdr,\controlvec_\text{cam},\textbf{r}_{\hat{x}})=u\cdot v_{\textbf{r}_{\hat{x}}}
\end{align}
\end{small}
$S_D$ denotes the stem layers of $D$ and $M_D$ maps the camera pose to the condition vector for the EG3D discriminator. We implement $T_G$ and $T_D$ using a ResNet architecture and zero-initialize both networks to ensure that the extra condition $\textbf{r}_{\hat{x}}$ blends into our existing deformable EG3D smoothly. 


 \begin{figure}[t]
    \centering
\scalebox{0.40}{
    \setlength{\tabcolsep}{0.5pt}
    \renewcommand{\arraystretch}{0.4}
  \begin{tabular}{ccccc}
      $\alpha=0$ & $\alpha=0.3$ & $\alpha=0.6$ & $\alpha=0.8$ & $\alpha=1$    \\ 
      \midrule 
    \includegraphics[valign=m,width=0.22\textwidth]{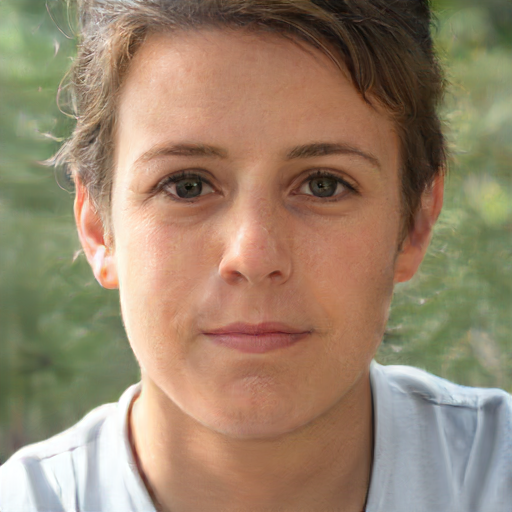} & 
    \includegraphics[valign=m,width=0.22\textwidth]{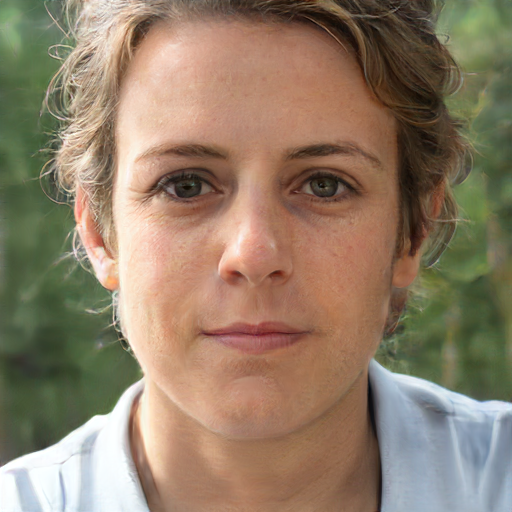} &  
    \includegraphics[valign=m,width=0.22\textwidth]{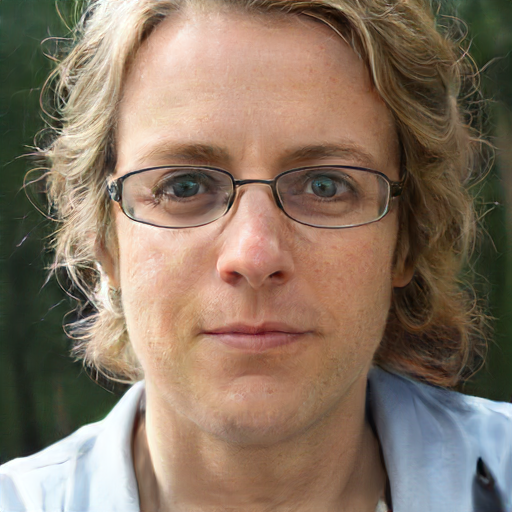} &
    \includegraphics[valign=m,width=0.22\textwidth]{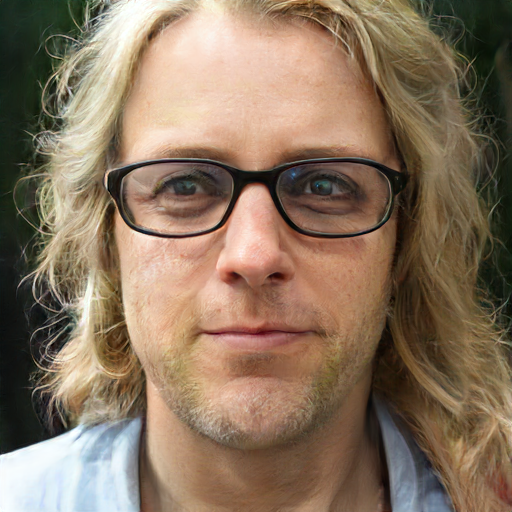} &
    \includegraphics[valign=m,width=0.22\textwidth]{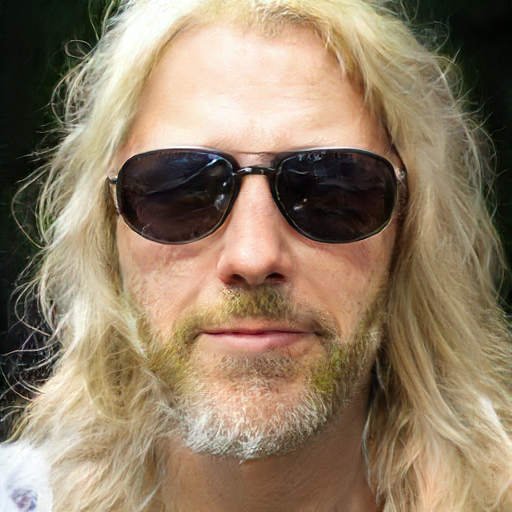}    \\
    
    
  \end{tabular}
    }
    \caption
    {
    \textbf{Increasing $\alpha$ increases prompt alignment.} \textit{Text Prompt: "Bearded man with long blond hair wearing glasses"} CLIP correlates regular glasses and sunglasses as increasing ``glasses'' intensity, and we observe a similar phenomenon.
    }    
    \label{fig:ablation_reg}
\end{figure}%


\begin{figure*}[h!]
    \centering
    \small
    \setlength{\tabcolsep}{1pt}

    \begin{minipage}{0.5\linewidth}
    \begin{tabular}{p{0.29\linewidth} cccc cccc cccc}
    \makecell[l]{\textbf{Input Image}} &
    \multicolumn{4}{c}{\includegraphics[valign=m,width=0.14\textwidth]{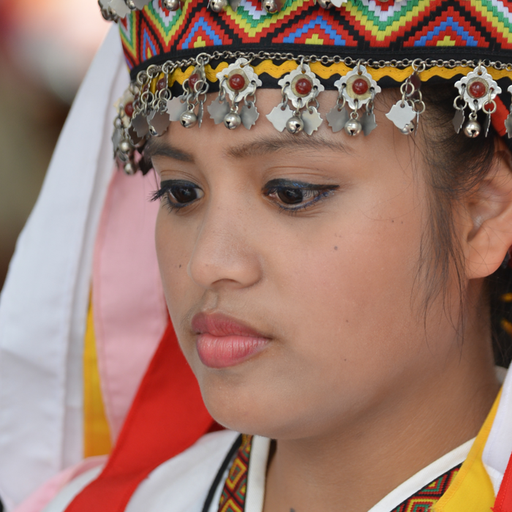}}    &
    \multicolumn{4}{c}{\includegraphics[valign=m,width=0.14\textwidth]{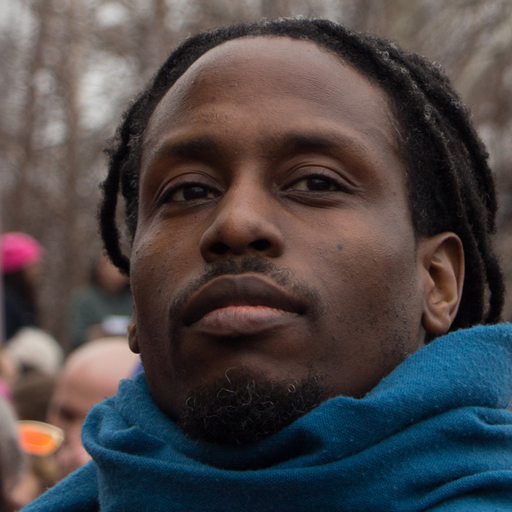}}   &
    \multicolumn{4}{c}{\includegraphics[valign=m,width=0.14\textwidth]{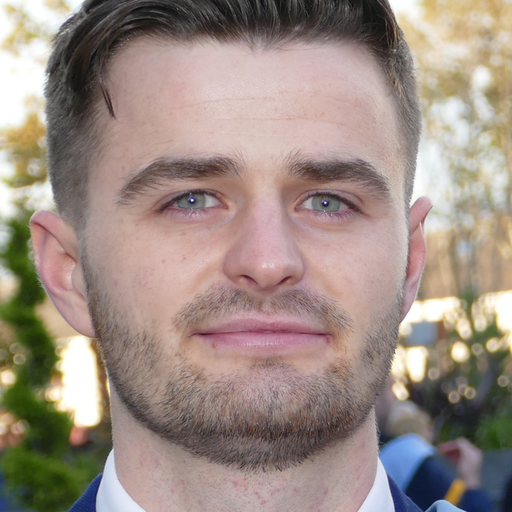}}   
    \\
    \rowcolor{LightGray}
    \makecell[l]{\textbf{Main prompt}} &
    \multicolumn{4}{c}{a close up of } &
    \multicolumn{4}{c}{a man with dreadlocks } &
    \multicolumn{4}{c}{a close up of a person}
    \\

    \rowcolor{LightGray}
     &
    \multicolumn{4}{c}{a person wearing a hat} &
    \multicolumn{4}{c}{wearing a blue scarf} &
    \multicolumn{4}{c}{ wearing a suit and tie }
    \\

    \makecell[l]{\textbf{Noise prompt}} &
    \multicolumn{4}{c}{ethnic group, canon  } &
    \multicolumn{4}{c}{in 2015, resistance, } &
    \multicolumn{4}{c}{bondi beach in the background,} 
    \\
    \makecell[l]{} &
    \multicolumn{4}{c}{ eos 1000d, shot on }  &
    \multicolumn{4}{c}{ quebec, black turtleneck,} &
    \multicolumn{4}{c}{ loosely cropped, red carpet}
    \\
    \makecell[l]{} &
    \multicolumn{4}{c}{ anamorphic lenses, round }  &
    \multicolumn{4}{c}{ mineral, exploitable image } &
    \multicolumn{4}{c}{ photography, post graduate, }
    \\
    \makecell[l]{} &
    \multicolumn{4}{c}{ dance, zun, emote, } &
    \multicolumn{4}{c}{ } &
    \multicolumn{4}{c}{uncompressed png, anxious   }
    \\
    \makecell[l]{} &
    \multicolumn{4}{c}{ colored vibrantly} &
    \multicolumn{4}{c}{ } &
    \multicolumn{4}{c}{steward of a new castle,}
    \\
    \makecell[l]{} &
    \multicolumn{4}{c}{} &
    \multicolumn{4}{c}{ } &
    \multicolumn{4}{c}{ultra high definition quality }
    \\

    \rowcolor{LightGray}
    \centering
    \makecell[l]{\textbf{Cosine similarity}} &
    \multicolumn{4}{c}{ 0.1838 } &
    \multicolumn{4}{c}{ 0.2551 } &
    \multicolumn{4}{c}{ 0.2273 }
    \\
    \rowcolor{LightGray}
    \centering
    \makecell[l]{\textbf{with main prompt}} &
    \multicolumn{4}{c}{  } &
    \multicolumn{4}{c}{  } &
    \multicolumn{4}{c}{  }

    \\
    \centering
    \makecell[l]{\textbf{Cosine similarity}} &
    \multicolumn{4}{c}{ 0.3367 } &
    \multicolumn{4}{c}{ 0.2701 } & 
    \multicolumn{4}{c}{ 0.3092 }
    \\
    \centering
    \makecell[l]{\textbf{with noise prompt}} &
    \multicolumn{4}{c}{  } &
    \multicolumn{4}{c}{  } &
    \multicolumn{4}{c}{  }

    \\
    \rowcolor{LightGray}
    \centering
    \makecell[l]{\textbf{Avg.~cosine sim.}} &
    \multicolumn{4}{c}{ 0.1221 } &
    \multicolumn{4}{c}{ 0.1248 } & 
    \multicolumn{4}{c}{ 0.1537 }
    \\
    \rowcolor{LightGray}
    \centering
    \makecell[l]{\textbf{with noise prompt}} &
    \multicolumn{4}{c}{  } &
    \multicolumn{4}{c}{  } &
    \multicolumn{4}{c}{  }
    \\
    \rowcolor{LightGray}
    \centering
    \makecell[l]{\textbf{across FFHQ}} &
    \multicolumn{4}{c}{  } &
    \multicolumn{4}{c}{  } &
    \multicolumn{4}{c}{  }

    \end{tabular}
    \end{minipage}
    \hspace{0.29\linewidth}
    \begin{minipage}{0.20\linewidth}
    \captionof{figure}
    {
    The comparison of cosine similarity between CLIP image embeddings of training images and CLIP text embeddings for prompts related to faces and those not related to faces. Noise prompts can have higher cosine similarity than face prompts. Further, the average cosine similarity between the noise prompts and the entire FFHQ dataset shows that $\textbf{r}_{\hat{x}_\text{noise}}$ is unique enough to identify \emph{each} corresponding training image. 
    }
    \label{fig:clipnoise}
    \end{minipage}
    \label{fig:supp_clip_noise}
    \vspace{-0.25cm}
\end{figure*}%

\subsection{Regularizing Noise in CLIP Embeddings}\label{sec:regularization}

\paragraph{Noise in CLIP.}
We verify its existence by showing the similarity between face-related \textit{main prompts} and \textit{noise prompts} that are not related to the face description when both are compared against a CLIP image embedding of the target image. To show this, we generate text of FFHQ training images using the third-party CLIP-Interrogator tool \footnote{https://huggingface.co/spaces/pharmapsychotic/CLIP-Interrogator}. Given the generated text prompts, we split them into two groups: face-related (main prompt) and face-unrelated (noise prompt); Figure~\ref{fig:clipnoise}. Then, we evaluate \textit{cosine} distance between CLIP image embedding and CLIP text embeddings. $\textbf{r}_{\hat{x}_\text{noise}}$ does exist and can outweigh the main facial appearance information in terms of cosine similarity. Furthermore, the average cosine similarity between the noise prompts and the entire FFHQ dataset shows that $\textbf{r}_{\hat{x}_\text{noise}}$ is unique enough to identify \emph{each} corresponding training image.

\paragraph{Distribution collapse from noise.}
This leads to distribution collapse during training of the proposed network. After disentangling 3D control from $\textbf{r}_{x}$, the remaining entanglement arises from $\textbf{r}_{\hat{x}_\text{noise}}$. As $\textbf{r}_{\hat{x}_\text{noise}}$ is highly specific to each $x$, each $x$ becomes identifiable solely from $\textbf{r}_{\hat{x}}$ and thus the conditional distribution $p(x\mid \textbf{r}_{\hat{x}})$ collapses to a delta distribution and all randomness of $x$ is lost once $\textbf{r}_{\hat{x}}$ is determined.

Since $G$ is trained to match $p(x\mid \textbf{r}_{\hat{x}})$, it is encouraged to be a deterministic function of $\textbf{r}_{\hat{x}}$ and completely ignore the source of randomness $\noisevec$, as sampling from $p(x\mid \textbf{r}_{\hat{x}})$ involves no randomness. This lack of randomness is highly undesirable for $G$: a generic prompt such as ``a blond person'' will be mapped to a single deterministic output rather than many diverse face images. This severely limits applications. Although we replace $\textbf{r}_{\hat{x}}$ with $\textbf{r}_t$ at inference time and $\textbf{r}_t$ contains no noise signal, the result will remain deterministic given that $G$ has learned to dissociate $\noisevec$ during training.

For this dissociation to happen, either $G$ has become a constant function w.r.t $\noisevec$, or $G$ is much more sensitive to the change of $\textbf{r}_{\hat{x}}$ than to the change of $\noisevec$. More formally:%
\begin{small}
\begin{align}
\frac{\partial G}{\partial \noisevec}=0  
\hspace{0.4cm} \label{eq:G_z}\\ \hspace{0.4cm} 
\left \| \frac{\partial G}{\partial \textbf{r}_{\hat{x}}}  \right \|_\text{F} \gg \left \| \frac{\partial G}{\partial \noisevec}  \right \|_\text{F}
\label{eq:G_z_bound}
\end{align}
\end{small}
where $\left \| \cdot \right \|_\text{F}$ denotes the Frobenius norm of the Jacobian. 

To address Eq.~\eqref{eq:G_z}, we force our model to retain the ability of unconditional generation by setting $\alpha=0$ with 50\% probability during training. For unconditional generation, $\noisevec$ is the only input to the volume synthesis process. By forcing the distribution of unconditional generation to match the training distribution $p(\mathcal{X})$, it is encouraged to produce diverse samples, which is directly at odds with $G$ being a constant function w.r.t $z$. 

For Eq.~\eqref{eq:G_z_bound}, a straightforward solution is to penalize $\left \| \partial G / \partial \textbf{r}_{\hat{x}}  \right \|_\text{F}$. However, this Jacobian term is too expensive to calculate. With the chain rule, we see that:
\begin{small}
\begin{align}
\left \| \frac{\partial G}{\partial \textbf{r}_{\hat{x}}}  \right \|_\text{F}=
\left \| \frac{\partial G}{\partial w_{\textbf{r}_{\hat{x}}}} \frac{\partial w_{\textbf{r}_{\hat{x}}}}{\partial \textbf{r}_{\hat{x}}}  \right \|_\text{F}\leqslant 
\left \| \frac{\partial G}{\partial w_{\textbf{r}_{\hat{x}}}} \right \|_2 \left \| \frac{\partial w_{\textbf{r}_{\hat{x}}}}{\partial \textbf{r}_{\hat{x}}}  \right \|_\text{F}
\end{align}
\end{small}
$\partial G/\partial w_{\textbf{r}_{\hat{x}}}$ is the computation bottleneck and not directly relevant to $\textbf{r}_{\hat{x}}$, we thus omit this Jacobian term and penalize $\left \| \partial w_{\textbf{r}_{\hat{x}}}/ \partial \textbf{r}_{\hat{x}}  \right \|_\text{F}$, which penalizes the upper bound of $\left \| \partial G / \partial \textbf{r}_{\hat{x}}  \right \|_\text{F}$. We formulate this penalty as a regularization term $R_{\textbf{r}_{\hat{x}}}$ and apply a stochastic approximator for efficient compute:
\begin{small}
\begin{align}
&R_{\textbf{r}_{\hat{x}}}= \left \| \frac{\partial w_{\textbf{r}_{\hat{x}}}}{\partial \textbf{r}_{\hat{x}}}  \right \|^2_\text{F}\\
&=\lim_{\sigma \to 0}\mathbb{E}_{\epsilon\sim\mathcal{N}(0,\sigma^2I)}\left[\frac{1}{\sigma^2}\left\| w_{\textbf{r}_{\hat{x}}+\epsilon}-w_{\textbf{r}_{\hat{x}}} \right\|^2 \right] \label{eq:jacob_reg} 
\end{align}
\end{small}
We show the proof of Eq.~\eqref{eq:jacob_reg} in the appendix. Additionally, we notice that the norm of $w_{\textbf{r}_{\hat{x}}}$ has the tendency to grow uncontrollably, driving conditional generation results to unrealistic regions. We penalize norm growth:%
\begin{small}
\begin{align}
R_\text{norm}=\left(\left\| w_{\textbf{r}_{\hat{x}}} \right\|-\left\| w \right\|\right)^2 \label{eq:norm_reg}
\end{align}
\end{small}
when deriving conditional style vector $w_{\textbf{r}_{\hat{x}}}$ from unconditional $w$ as in \eqref{eq:alpha_weight}.


 \begin{figure}[t]
    \centering
    \small
\scalebox{0.45}{
    \setlength{\tabcolsep}{1.5pt}
    \renewcommand{\arraystretch}{0.4}
  \begin{tabular}{p{2.5cm} ccc | ccc }
     {\bf Text Prompt} & \multicolumn{3}{
c}{\bf Full Model} &  \multicolumn{3}{c}{\bf Without Noise Regularization }   \\ 
    \makecell[l]{This man has a\\big beard and\\black hair.} &
    \includegraphics[valign=m,width=0.14\textwidth]{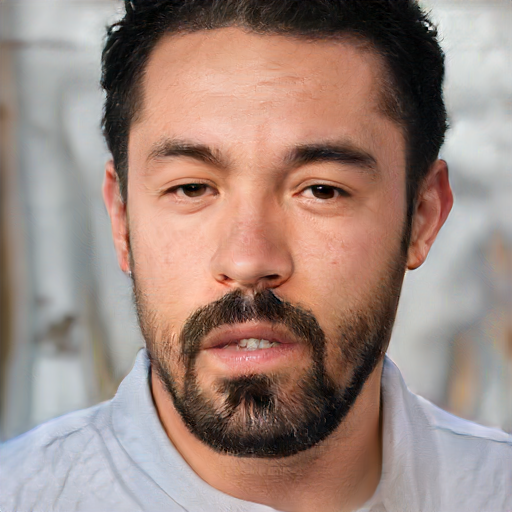} & 
    \includegraphics[valign=m,width=0.14\textwidth]{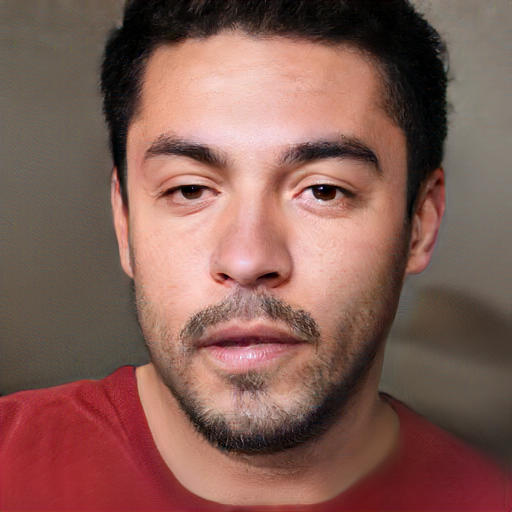} &  
    \includegraphics[valign=m,width=0.14\textwidth]{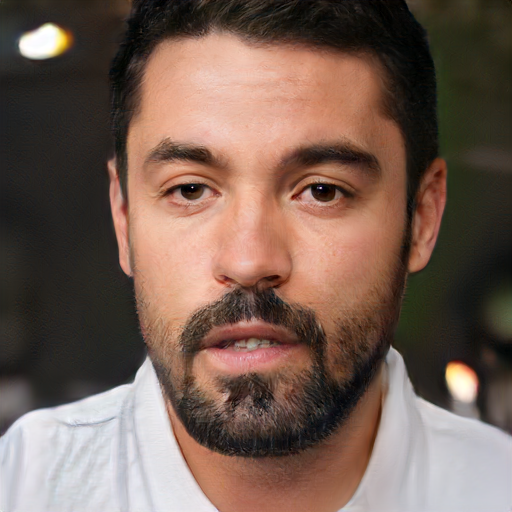} &
    \includegraphics[valign=m,width=0.14\textwidth]{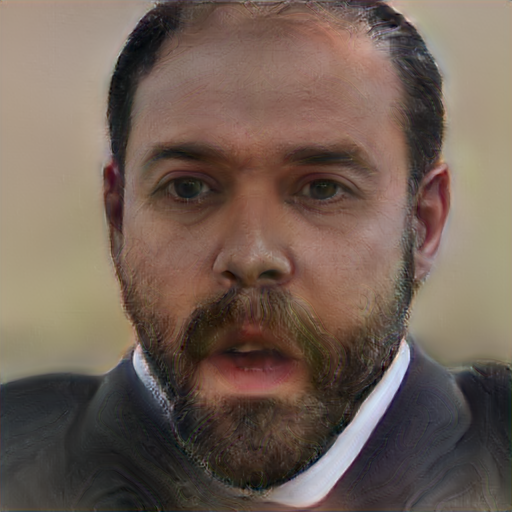} &
    \includegraphics[valign=m,width=0.14\textwidth]{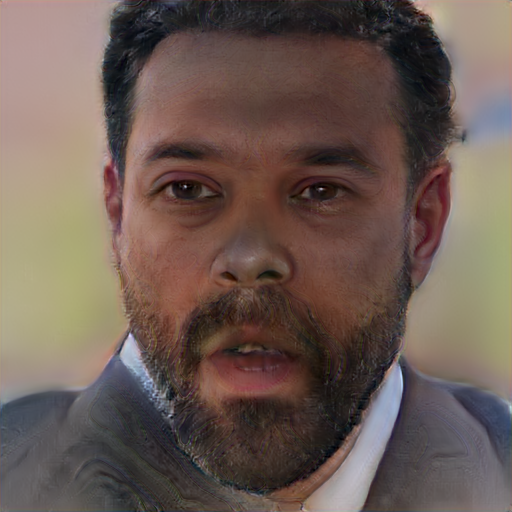} &
    \includegraphics[valign=m,width=0.14\textwidth]{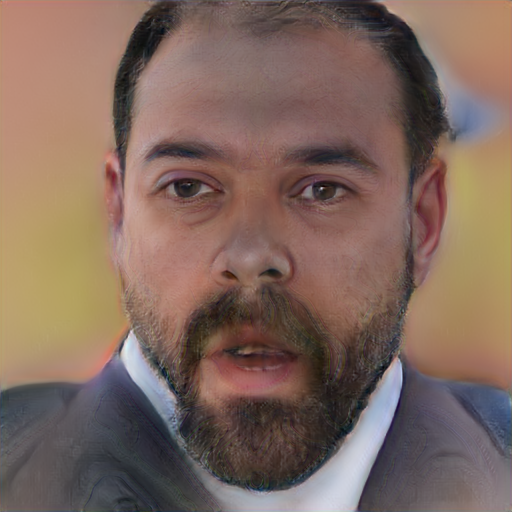} \\
    
    
    \makecell[l]{This young man\\has bangs and\\black hair.} &
    \includegraphics[valign=m,width=0.14\textwidth]{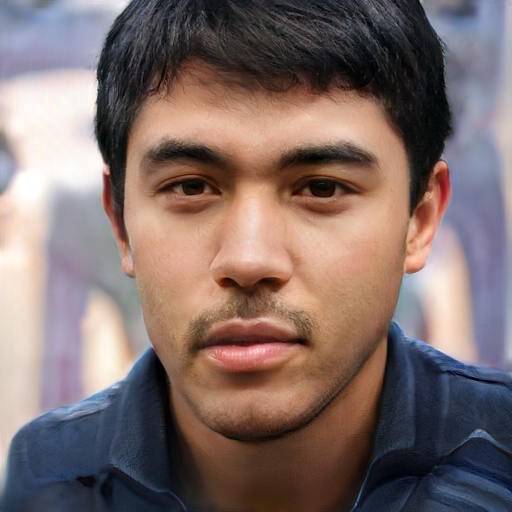} & 
    \includegraphics[valign=m,width=0.14\textwidth]{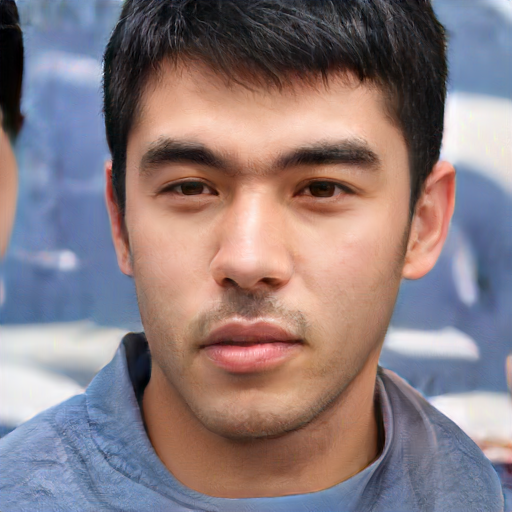} &  
    \includegraphics[valign=m,width=0.14\textwidth]{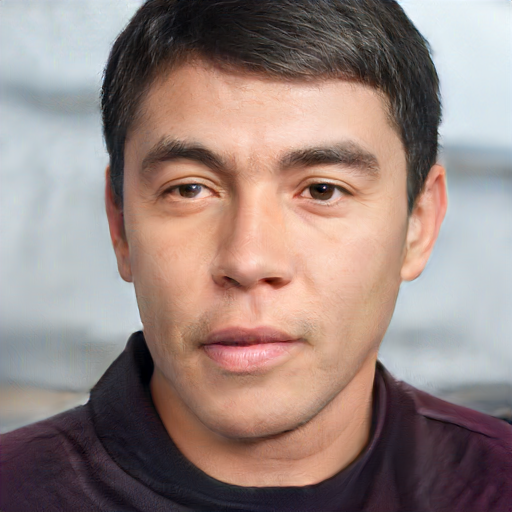} &
    \includegraphics[valign=m,width=0.14\textwidth]{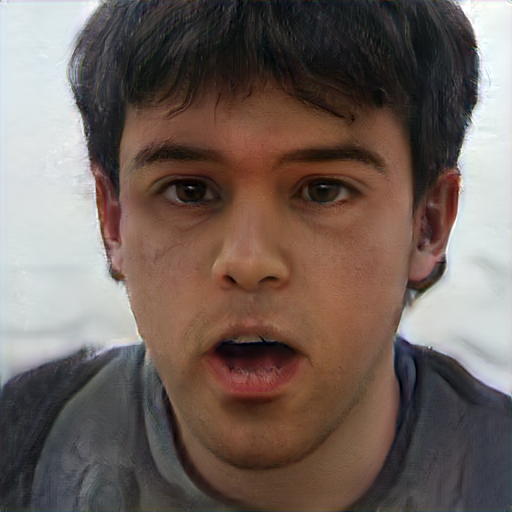} &
    \includegraphics[valign=m,width=0.14\textwidth]{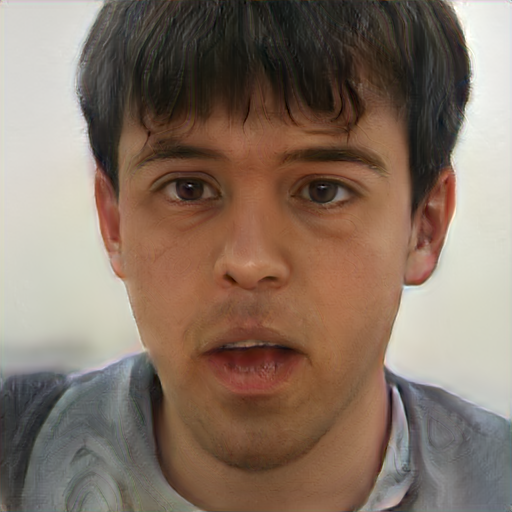} &
    \includegraphics[valign=m,width=0.14\textwidth]{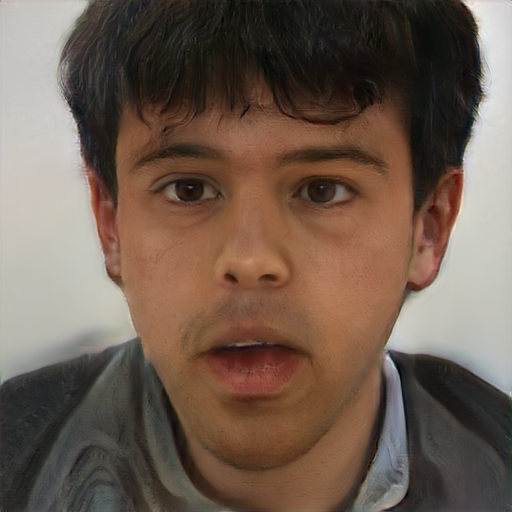} 

  \end{tabular}
    }
    \caption
    {
    \textbf{Showing distribution collapse.} Our noise regularization improves diversity and quality. [FFHQ 5mil.~images.]
    }    
    \label{fig:ablation_reg}
\end{figure}%

With our regularization techniques, we solve the distribution collapse problem allowing high-quality and diverse generation. Figure \ref{fig:ablation_reg} verifies our hypothesis that the model cannot help but dissociate the source of randomness $z$ under a stronger (text-)conditioning signal without the proposed Jacobian regularization. With both regularizers in place, we trade off a slight compromise in text-alignment overall (`bangs'). As a side control, we also provide a smooth transition from unconditional generation to conditional generation by manipulating $\alpha$, allowing the user to trade-off between alignment and fidelity (Fig.~\ref{fig:blue_eyes}).

 \begin{figure*}[t]
    \centering
    \small
\scalebox{0.5}{
    \setlength{\tabcolsep}{1pt}
    \renewcommand{\arraystretch}{0.4}
  \begin{tabular}{c c | c c | c c | c c | c c}
    \multicolumn{2}{l}{She has black hair and wears eyeglasses.} &
    \multicolumn{2}{l}{This woman has bushy eyebrows and straight hair.} &
    \multicolumn{2}{l}{He has a long brown beard.} &
    \multicolumn{2}{l}{She has straight brown hair and is wearing lipstick.} &
    \multicolumn{2}{l}{Bearded man with long blond hair wearing glasses.}
    \\
    \includegraphics[valign=m,width=0.16\textwidth]{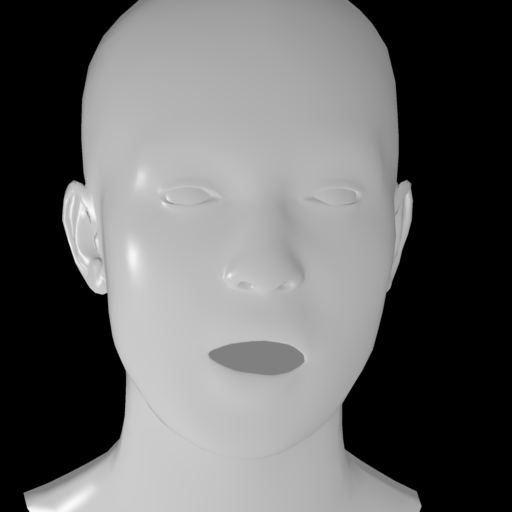} &
    \includegraphics[valign=m,width=0.16\textwidth]{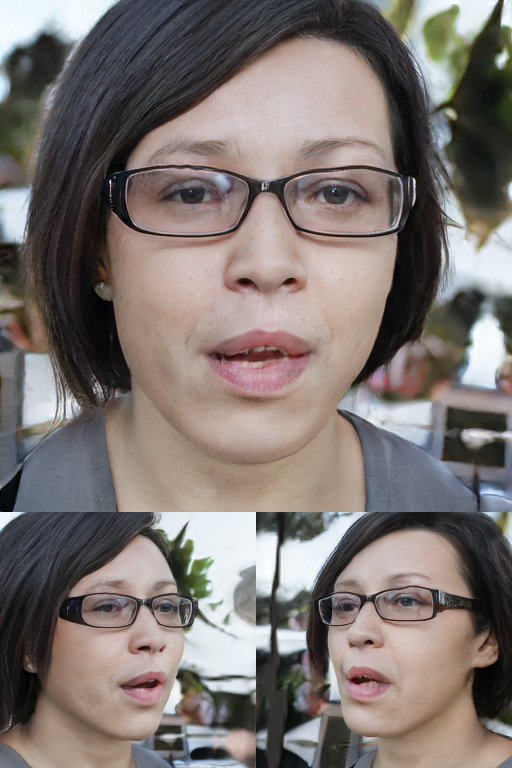} 
    &
    \includegraphics[valign=m,width=0.16\textwidth]{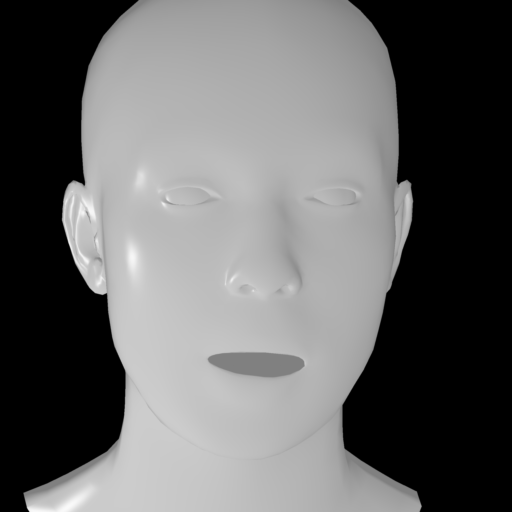} &
    \includegraphics[valign=m,width=0.16\textwidth]{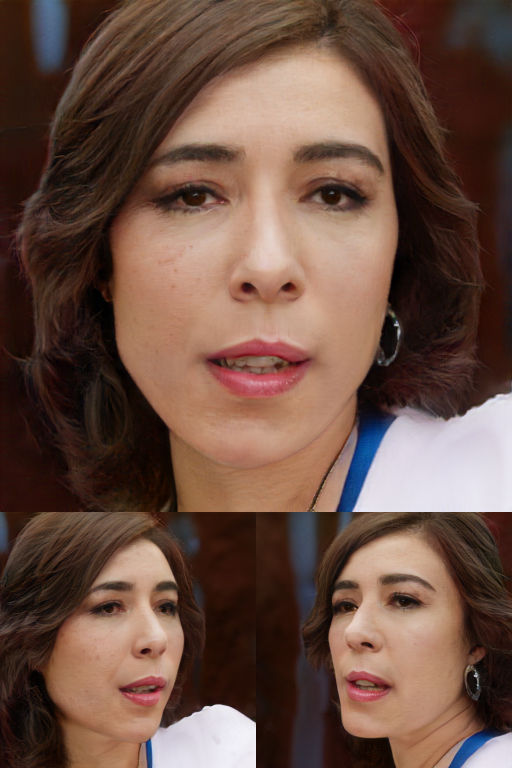} 
    &
    \includegraphics[valign=m,width=0.16\textwidth]{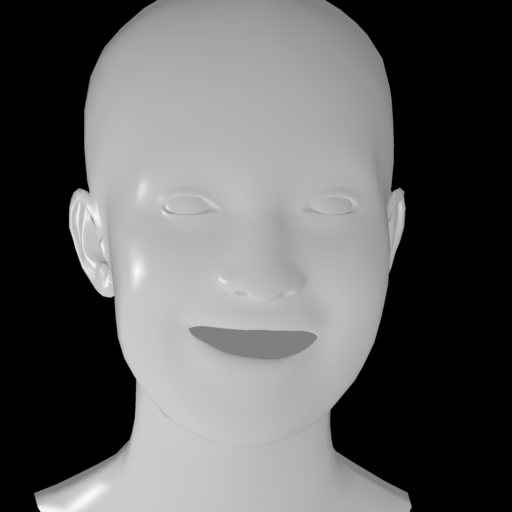} &
    \includegraphics[valign=m,width=0.16\textwidth]{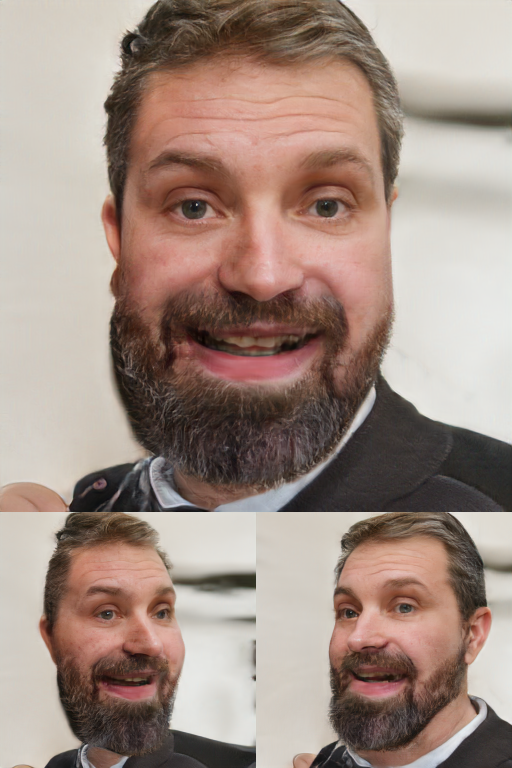}
    &
    \includegraphics[valign=m,width=0.16\textwidth]{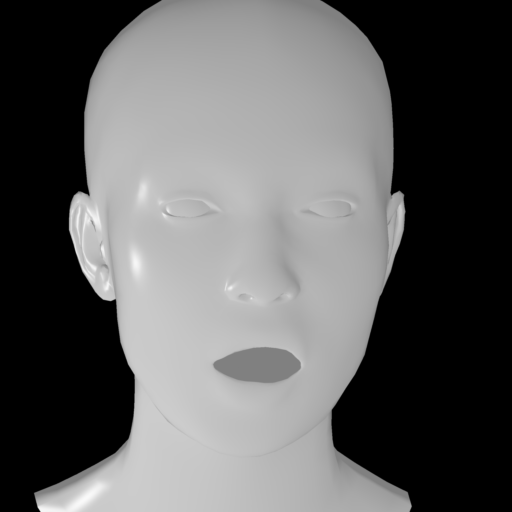} &
    \includegraphics[valign=m,width=0.16\textwidth]{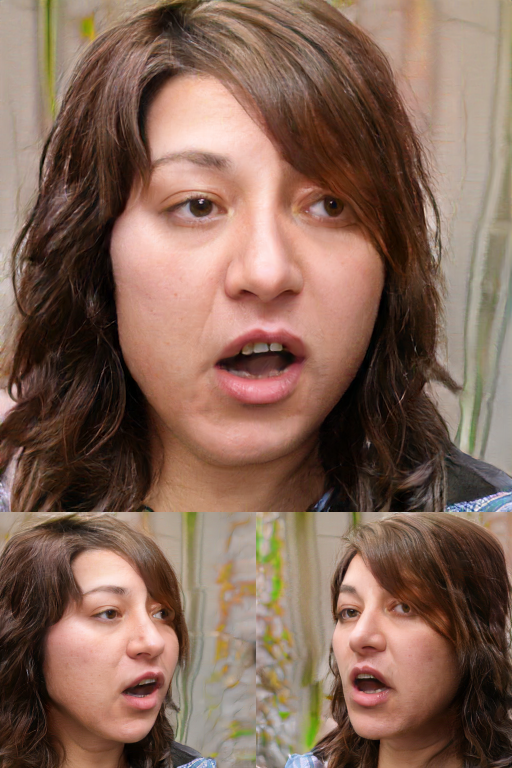}
    &
    \includegraphics[valign=m,width=0.16\textwidth]{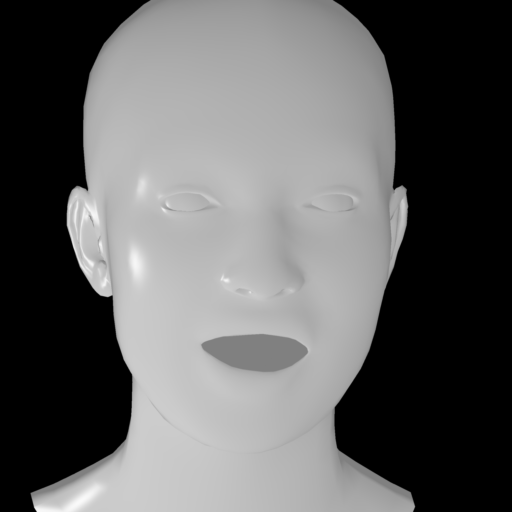} &
    \includegraphics[valign=m,width=0.16\textwidth]{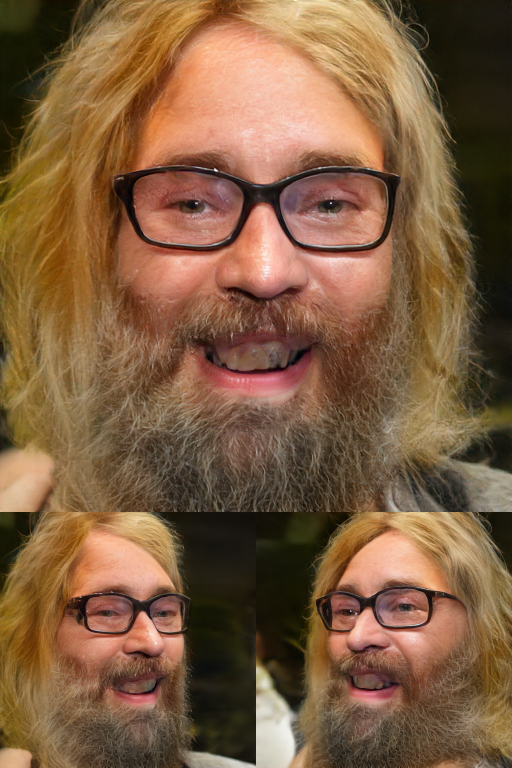}
    

  \end{tabular}
    }
    \caption
    {
    \textbf{Qualitative Evaluation} On text-to-3D portrait generation with explicit geometry control.}
    \label{fig:method_eval}
\end{figure*}%

\section{Experiments}
\label{sec:results}




\noindent \textbf{Datasets.}
\label{subsec:datasets}
We use FFHQ \cite{karras2019style} and Multi Modal CelebA-HQ (MMCelebA) \cite{xia2021tedigan} at 512$\times$512 resolution following prior work. While MMCelebA contains text annotations, we do not use any when training our model, and the data processing procedure follows exactly as in FFHQ.

We augment both datasets with horizontal flips and estimate the camera parameters for each image using an off-the-shelf model following EG3D. For FLAME parameter estimation, we adopt DECA to obtain initial results. The DECA estimates are not directly applicable due to the camera model differences between DECA (orthographic) and EG3D (perspective). We further optimize the initial estimates from DECA using a projected facial landmark loss to reconcile this difference. Finally, we optimize the scale and translation of the FLAME mesh to match the cropping of EG3D; other mesh postprocessing steps such as water-tightening and simplification follow GNARF.


\noindent \textbf{Model and Optimization.} For the EG3D backbone, we initialize the generator weights from the Egger et al.~public checkpoint on FFHQ and largely follow their training routine and losses, including the non-saturating adversarial loss, R1 gradient penalty, and density regularization. However, we remove generator pose conditioning and lower learning rate $\gamma=0.001$ for both generator $G$ and discriminator $D$. For faster deformation computation, we simplify the FLAME mesh to 2500 triangles, but use the full mesh for generating the mesh render $rdr$ in discriminator conditioning. Lastly, we add our regularizations $R_{c_x}$ and $R_\text{norm}$ to the training objective, and empirically set their weights to 0.01 and 10.

\noindent \textbf{Metrics.}
For image generation quality, we use \textit{Frechet Inception Distance (FID)} \cite{FID_heusel2017gans} and \textit{Kernel Inception Distance (KID)} \cite{KID_binkowski2018demystifying}. For semantic consistency, we use \textit{CLIP score}, which is the \textit{cosine} similarity between a CLIP image embedding and a CLIP text embedding. 


 \begin{figure*}[t]
    \centering
    \small
\scalebox{0.62}{
    \setlength{\tabcolsep}{1pt}
    \renewcommand{\arraystretch}{0.4}
  \begin{tabular}{cc|ccccc}
    TG-3DFace---Original & 
    TG-3DFace---``Blue eyes'' & 
    $\alpha=0$ & $\alpha=0.3$ & $\alpha=0.6$ & $\alpha=0.8$ & $\alpha=1$  \\ 

    \includegraphics[valign=m,width=0.2\textwidth]{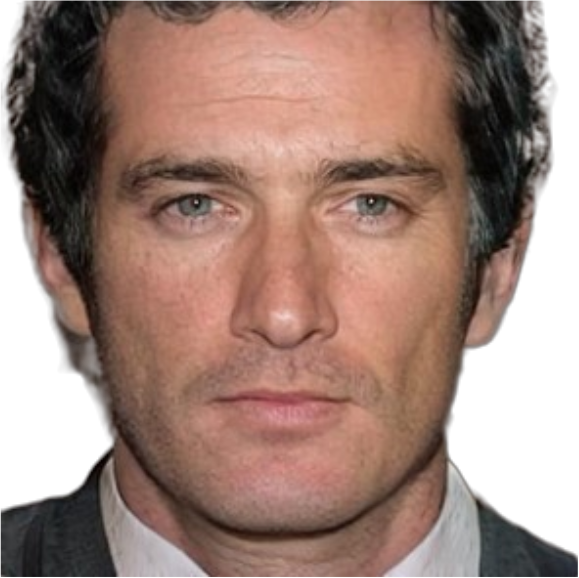} &
    \includegraphics[valign=m,width=0.2\textwidth]{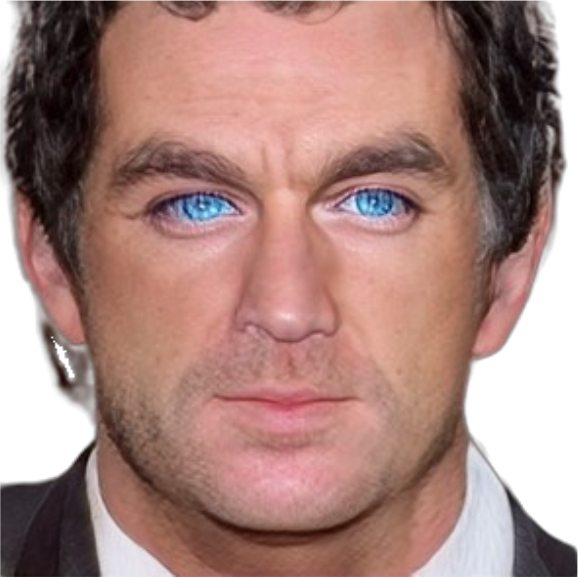} &
    \includegraphics[valign=m,width=0.2\textwidth]{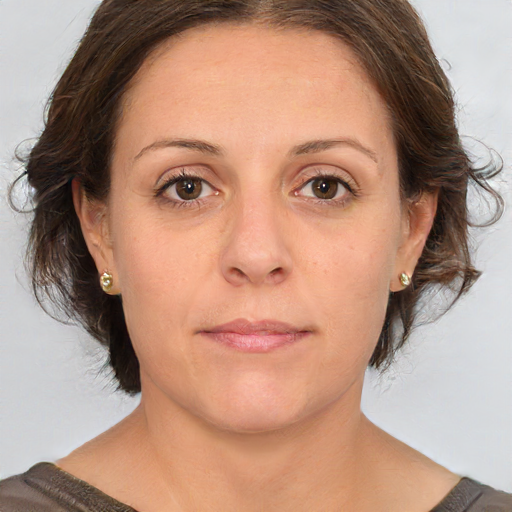} & 
    \includegraphics[valign=m,width=0.2\textwidth]{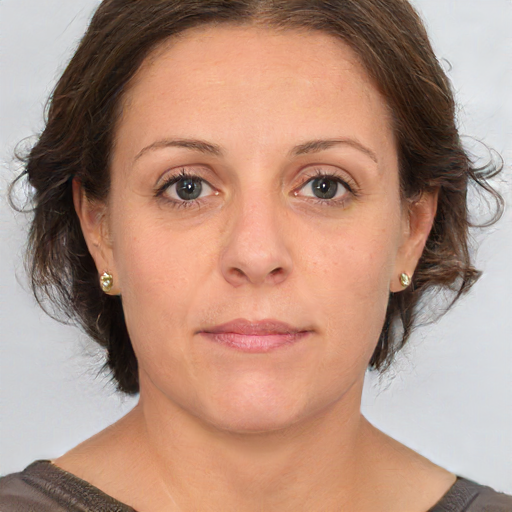} &  
    \includegraphics[valign=m,width=0.2\textwidth]{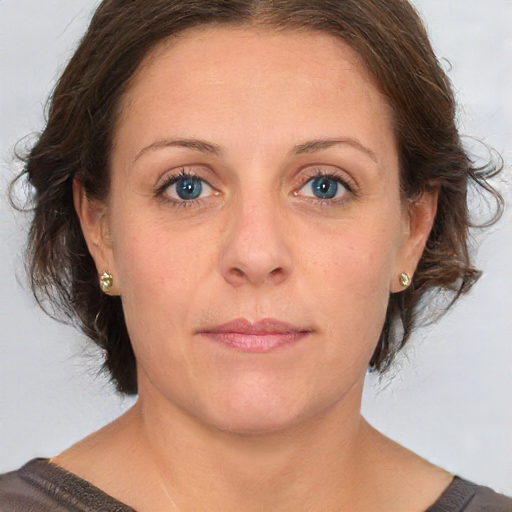} &
    \includegraphics[valign=m,width=0.2\textwidth]{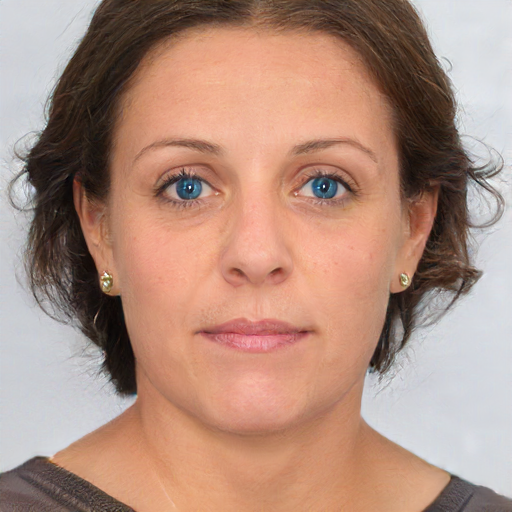} &
    \includegraphics[valign=m,width=0.2\textwidth]{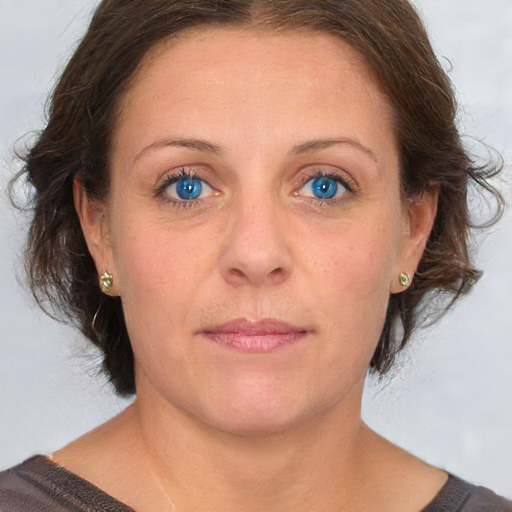}

  \end{tabular}
    }
    \vspace{-0.2cm}
    \caption
    {
    Text-guided 3D face appearance manipulation. Increasing $\alpha$ increases prompt alignment, which is not possible in TG-3DFace \cite{tg_3dface_yu2023towards}. As there is no public TG-3DFace~\cite{tg_3dface_yu2023towards} code, we took images from their original paper and used the same ``blue eyes'' prompt.
    }    
    \label{fig:blue_eyes}
\end{figure*}%
 \begin{figure}[t]
    \centering
    \small
\scalebox{0.55}{
    \setlength{\tabcolsep}{1pt}
    \renewcommand{\arraystretch}{0.4}
  \begin{tabular}{p{2.5cm} cc cc c}
     {\bf Text Prompt} & \multicolumn{2}{
c}{\bf CLIPortrait (Ours)} &  \multicolumn{2}{c}{\bf CLIPFace \cite{aneja2023clipface}} &  \multicolumn{1}{c}{\bf TG3DFace \cite{tg_3dface_yu2023towards}}  \\ 
    \makecell[l]{This woman has\\arched eyebrows\\and wavy hair.\\She is wearing\\red lipstick.} &
    \includegraphics[valign=m,width=0.14\textwidth]{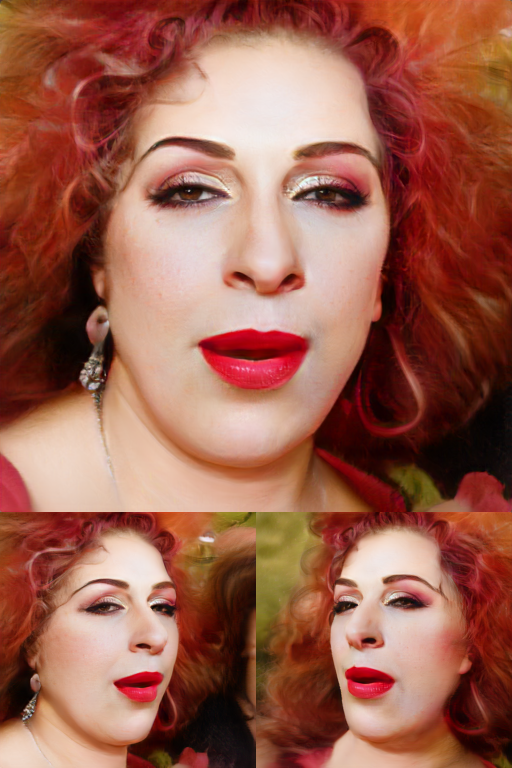} & 
    \includegraphics[valign=m,width=0.14\textwidth]{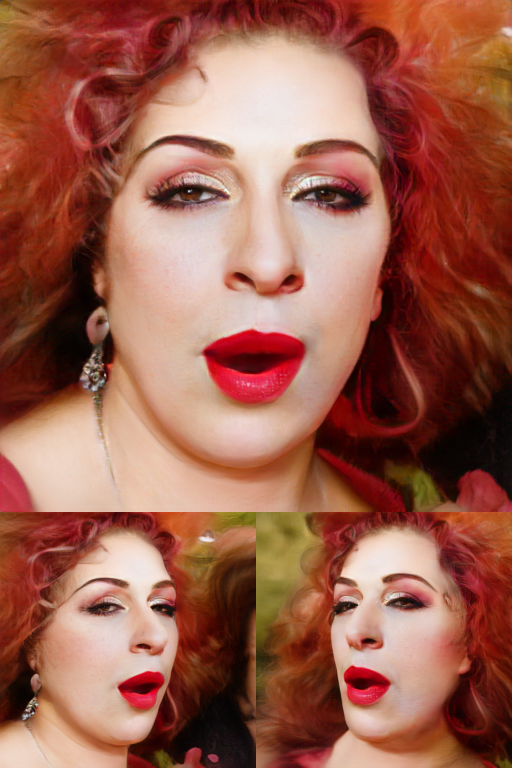}  &
    \includegraphics[valign=m,width=0.14\textwidth]{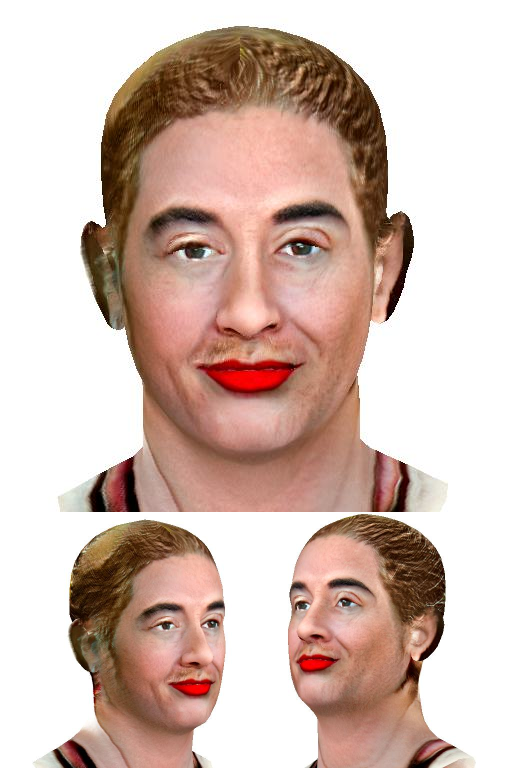} &
    \includegraphics[valign=m,width=0.14\textwidth]{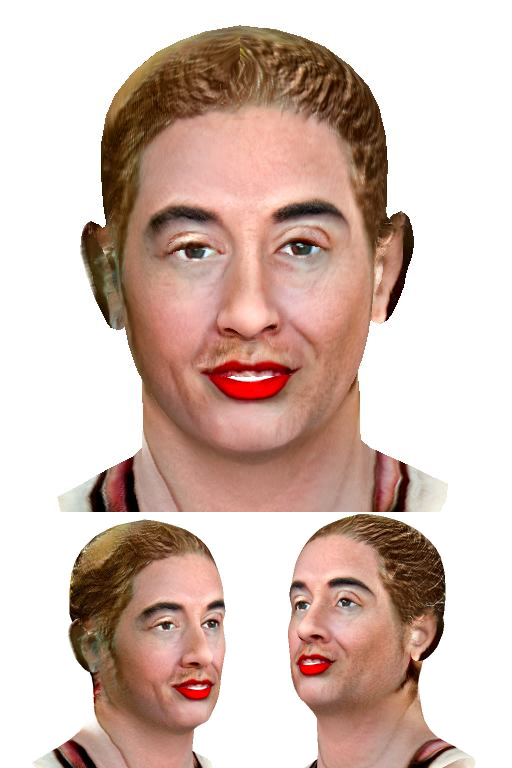} &
    \includegraphics[valign=m,width=0.14\textwidth]{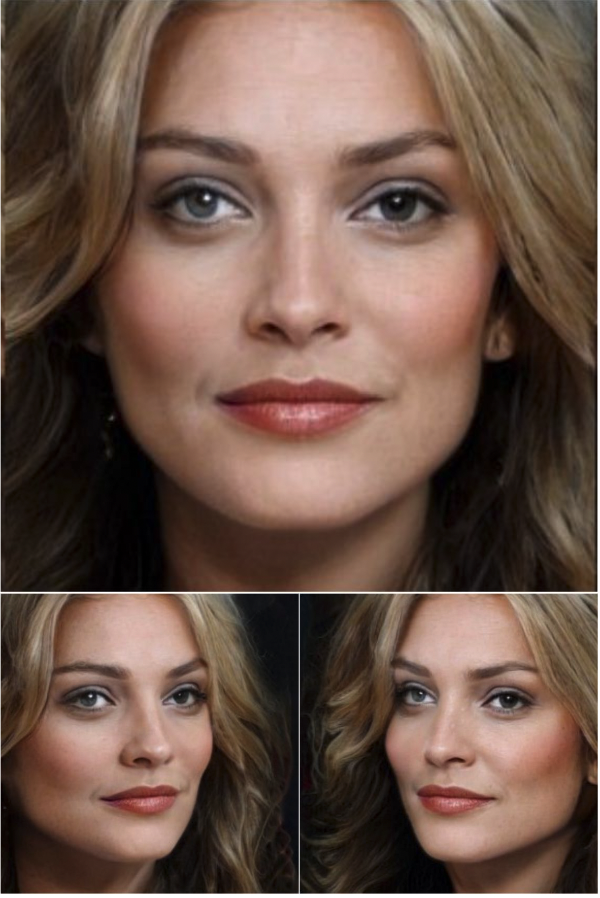}   \\
    
      \\

    
    \makecell[l]{This woman has\\blonde hair\\and pale skin.} &
    \includegraphics[valign=m,width=0.14\textwidth]{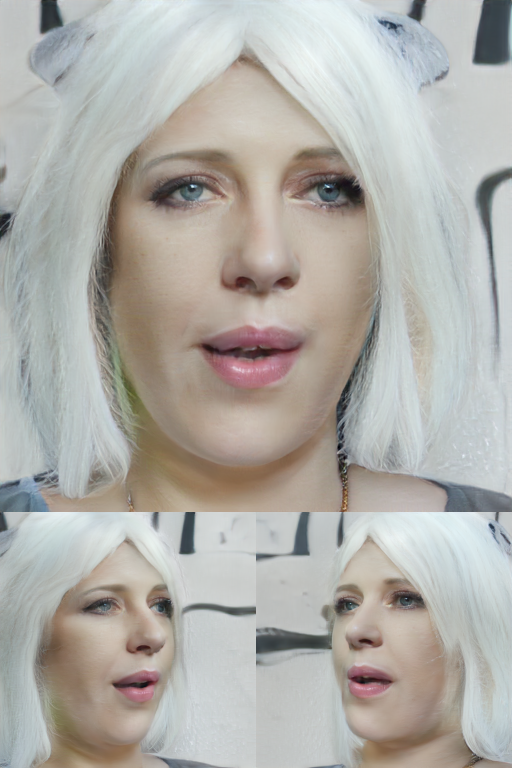} & 
    \includegraphics[valign=m,width=0.14\textwidth]{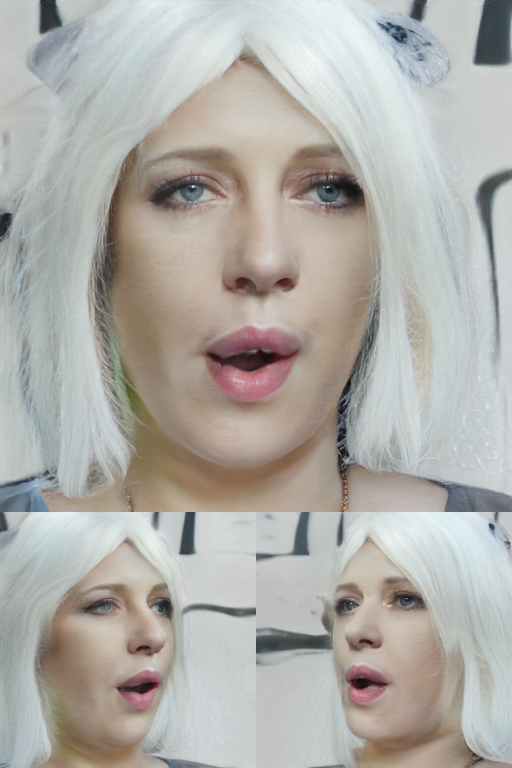} &  
    \includegraphics[valign=m,width=0.14\textwidth]{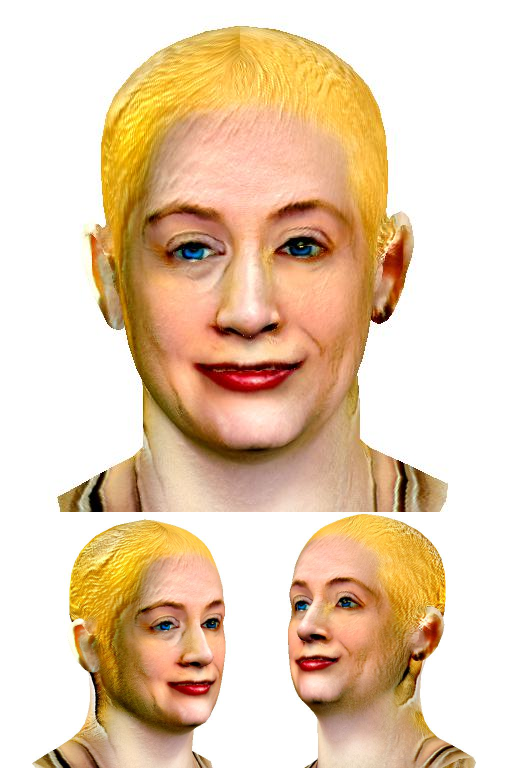} &
    \includegraphics[valign=m,width=0.14\textwidth]{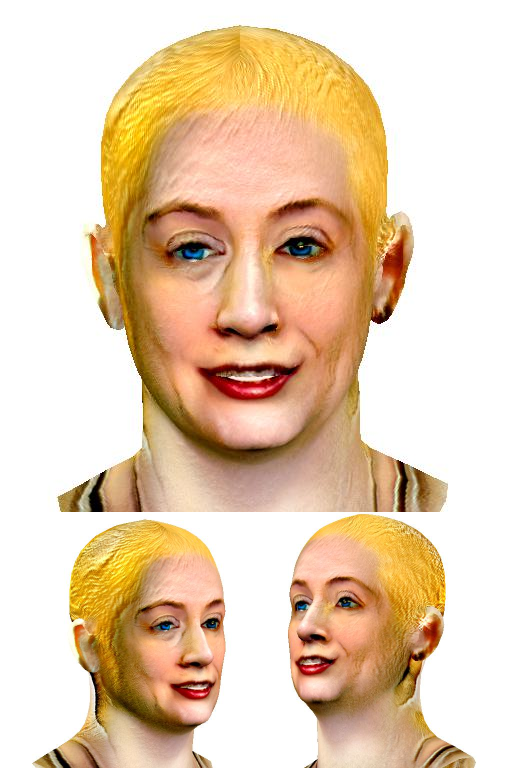} &
    \includegraphics[valign=m,width=0.14\textwidth]{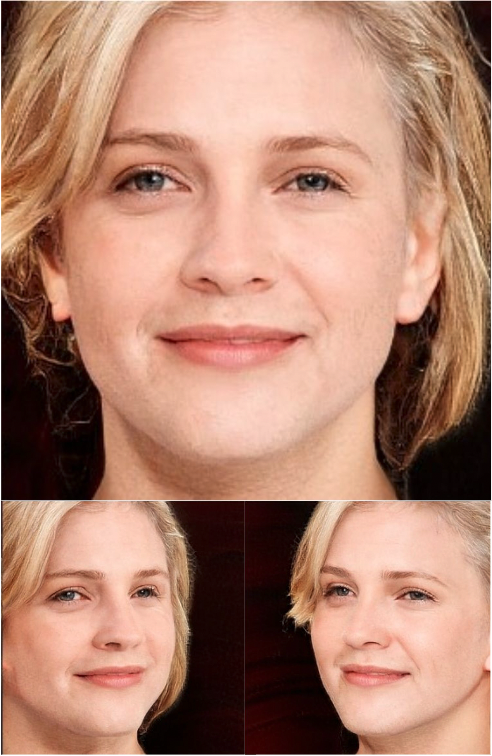}   \\
    

  \end{tabular}
    }
    \caption
    {
    \textbf{Text and 3D control (face shape, expression, and camera).} Our method shows improved quality against CLIPFace \cite{aneja2023clipface}, and improved diversity and control against TG3DFace \cite{tg_3dface_yu2023towards}, which cannot vary face shape or expression. These examples have $\alpha=1$, showing the most strong or dramatic response to the text prompt, e.g., very red lips or very blonde (white) hair.
    }    
    \label{fig:methods_comparison}
\end{figure}%




\subsection{Comparison on Text-to-3D Face Generation}
\label{subsec:comparison_textto3D}
We evaluate the performance of CLIPortrait on open-vocabulary text against the texture-mapping-based CLIPFace method~\cite{aneja2023clipface} and generative TG-3DFace method~\cite{tg_3dface_yu2023towards}. In qualitative comparisons (Fig.~\ref{fig:methods_comparison},~\ref{fig:method_eval}), we see improved image quality against CLIPFace and improved diversity and control against TG-3DFace. For quantitative evaluation, we compare with six additional text-guided 3D face generation methods (Tab.~\ref{tab:methods_comp_all}). On FFHQ~\cite{karras2019style}, 
we evaluate the reality and diversity of rendered images by computing FID and KID scores using random samples for noise and CLIP embedding. On MMCelebA \cite{xia2021tedigan}, we follow the same random sampling as FFHQ for FID score computation and we use provided text annotations for \textit{CLIP score} computation. For experiment fairness, we use the given 3DMM coefficients associated with the samples. On both datasets, our method demonstrates higher fidelity and semantic consistency.

\begin{table}[t]
    \centering
    \setlength{\tabcolsep}{2mm}{
    \vspace{5pt}
    \renewcommand\arraystretch{1.0}
    \resizebox{1.0\linewidth}{!}{
    \begin{tabular}{lcc|lcc}
    \toprule
    \multirow{2}{*}{Method}  & 
    \multicolumn{2}{c}{\textbf{FFHQ}} & \multirow{2}{*}{Method}  & \multicolumn{2}{c}{\textbf{MM-Celeb-A-HQ}}\\ 
    & {FID\;$\downarrow$} &  {KID\;$\downarrow$} & & {FID\;$\downarrow$} &  {CLIP Score\;$\uparrow$} \\  
    \midrule
    Text2Mesh* \cite{michel2022text2mesh}  & 219.59  &  0.185   & SEA-T2F \cite{sun2021multi}  &  93.8  &  20.8   \\ 
    ClipMatrix* \cite{jetchev2021clipmatrix}  & 198.34  &  0.180  & ControlGAN \cite{li2019controllable}  &  74.5  & 21.3   \\ 
    FlameTex* \cite{FLAME:SiggraphAsia2017}  & 88.95  &  0.053   & AttnGAN \cite{xu2018attngan}  &  51.6  &  21.5   \\ 
    CLIPFace* \cite{aneja2023clipface} &   80.3  &  0.032   & TG-3DFace \cite{tg_3dface_yu2023towards}  &    39.0  &   22.7   \\
    CLIPortrait & \bf 5.87  &  \bf 0.00217   & CLIPortrait  &  \bf 31.8    &  \bf 23.1   \\ 
    \bottomrule
    \end{tabular}}}
    \caption{Quantitative comparison. See Fig.~\ref{fig:methods_comparison} for a qualitative comparison. *: These models are rasterization-based and cannot achieve the same level of fidelity as volume rendered models.} 
    \label{tab:methods_comp_all}
    \vspace{-15pt}
\end{table}


\noindent \textbf{Text-Image Matching Performance.} However, CLIP score is not a reliable measure on its own without FID. In Table \ref{tab:methods_comp_all}, CLIPortrait achieves good scores in both. But, in a comparison just of CLIP score, the CLIPFace method beats CLIPortrait on the FFHQ-Text dataset~\cite{ffhq_text_zhou2021generative} (Tab.~\ref{tab:methods_comp_clipscore}). Since the authors do not provide the text prompts used in the original CLIPFace publication, we used the FFHQ-Text dataset to compute CLIP score. While it is higher, even distant inspection shows that samples created by CLIPFace do not look realistic. In contrast, our approach produces plausible appearance for the face as well as details in the hair or the presence of eyeglasses.


\begin{table}[t]
    \centering
    \setlength{\tabcolsep}{1mm}{
    \vspace{5pt}
    \renewcommand\arraystretch{1.0}
    \resizebox{1.0\linewidth}{!}{
    \begin{tabular}{lcc cccc} 
    \toprule
    \multirow{2}{*}{\bf Method} & \multirow{2}{*}{\bf CLIP Score} & \multirow{2}{*}{\bf Generation Time} &  \multicolumn{4}{c}{ \bf Text Prompts: } \\
    & & & \bf (a) & \bf (b) & \bf (c) &  \bf (d)  \\
    \midrule
    CLIPFace \cite{aneja2023clipface} &   24.1  &  24 minutes &   \includegraphics[valign=m,width=0.14\textwidth]{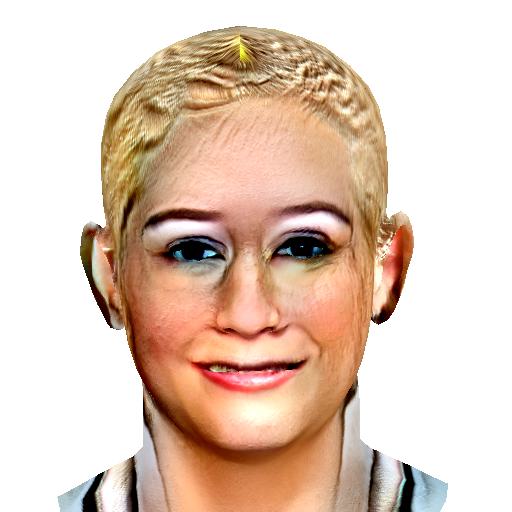} &   \includegraphics[valign=m,width=0.14\textwidth]{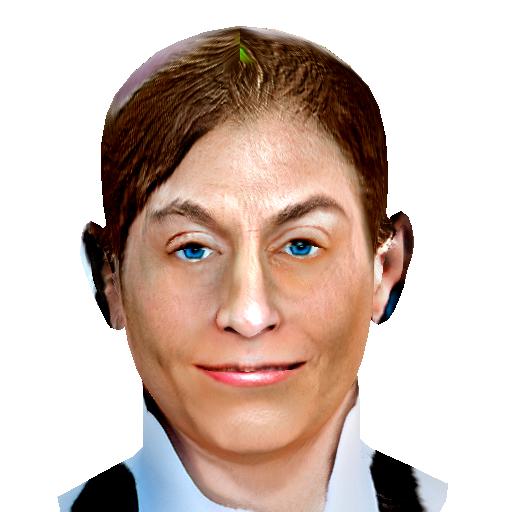} &   \includegraphics[valign=m,width=0.14\textwidth]{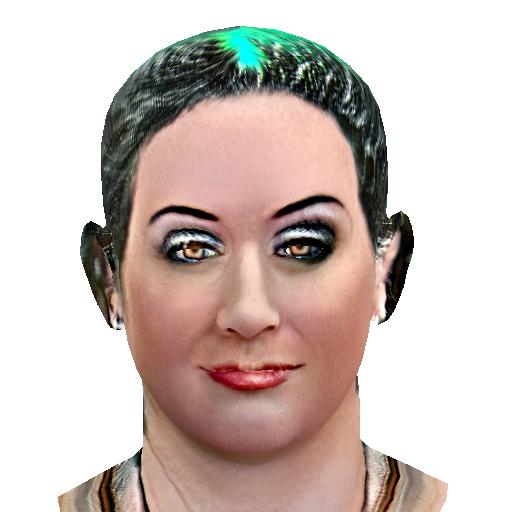} &   \includegraphics[valign=m,width=0.14\textwidth]{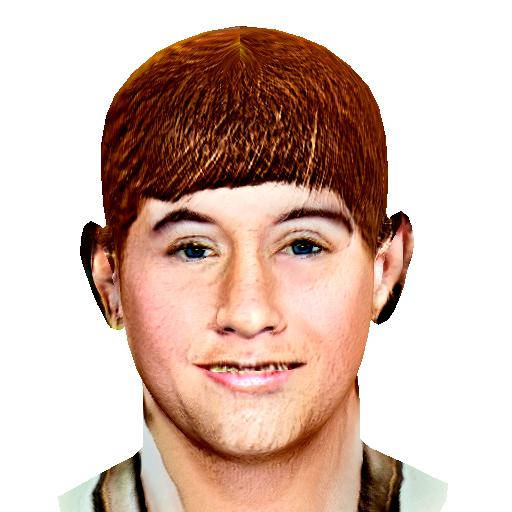}  \\
    CLIPortrait & 22.3  &  0.10 seconds  &  \includegraphics[valign=m,width=0.14\textwidth]{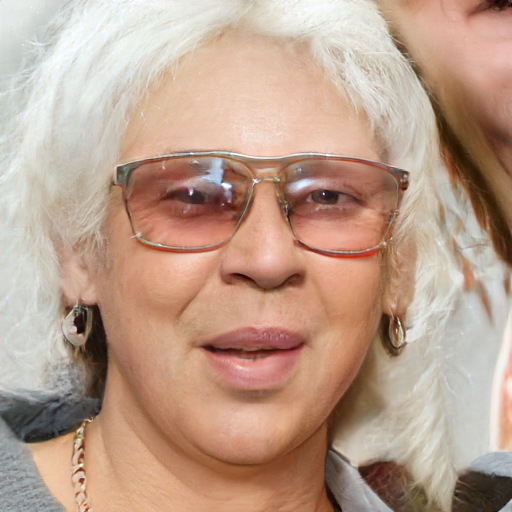} &  \includegraphics[valign=m,width=0.14\textwidth]{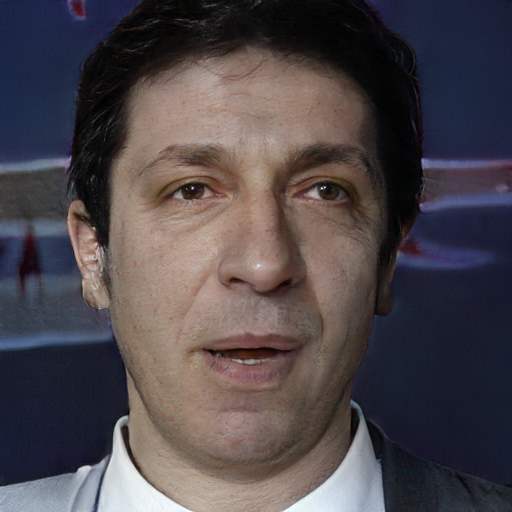} &  \includegraphics[valign=m,width=0.14\textwidth]{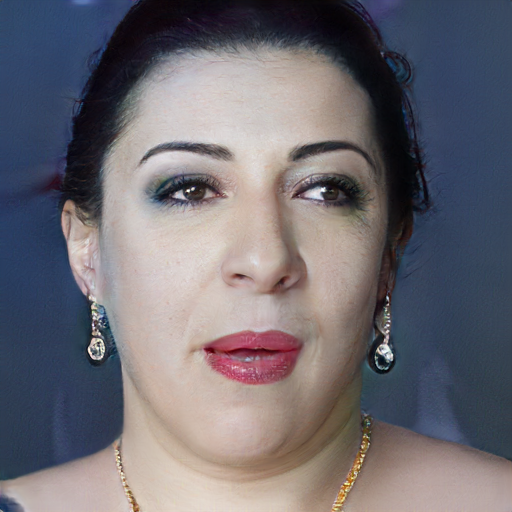} &  \includegraphics[valign=m,width=0.14\textwidth]{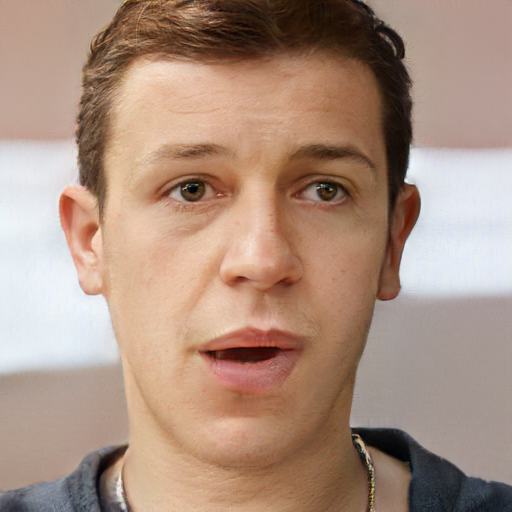}  \\ 
    \bottomrule
    \end{tabular}}}
    \caption{\textbf{CLIP score is a poor indicator of quality.} Our approach CLIPortrait generates more plausible portraits than CLIPFace. Text prompts: \textbf{(a)} ``She has blonde hair and wears eyeglasses'' \textbf{(b)} ``He has straight hair, and bags under eyes. He is wearing a necktie'', \textbf{(c)} ``The person wears heavy makeup, necklace. She has arched eyebrows, and black hair'', \textbf{(d)} ``This young person has brown hair''}
    \vspace{-5pt}
    \label{tab:methods_comp_clipscore}
\end{table}

\noindent \textbf{Inference Time Efficiency.} We present the generation speed of our method against CLIPFace in Table \ref{tab:methods_comp_clipscore}. The metrics were measured from the moment a new text prompt was inputted until the generation of a 3D portrait. Optimization-based text-to-3D face method CLIPFace \cite{aneja2023clipface} requires training from scratch for every new text prompt, resulting in a minimum time cost of $24$ minutes. In comparison, \sysname achieves generation speed of approximately $0.10$ seconds in the same settings with CLIPFace \cite{aneja2023clipface}. These results demonstrate the efficiency of our method in generating high-quality 3D-aware portraits from text prompts at smooth, nearly interactive frame rates.

\vspace{-5pt}
\section{Related Work}
\label{sec:rel_work}

\newcommand\rot[1]{\rlap{\rotatebox{45}{#1}}}
\newcommand\OK{$\checkmark$}

\begin{table}[t]
\setlength\tabcolsep{5pt} 
\centering
    \resizebox{0.8\linewidth}{!}{
     \begin{tabular}{@{} *{8}{l} }
     
      \footnotesize{\rot{Text  guidance}} &
      \footnotesize{\rot{3D view consistency}} &
      \footnotesize{\rot{3D geometry control}} &
      \footnotesize{\rot{Appearance editing}} &
      \footnotesize{\rot{Image only training}} &
      \footnotesize{\rot{Photorealistic render}} &
     \footnotesize{Method} \\
     \midrule 
         &  \OK  &     &     & \OK & \OK &  \scriptsize{EG3D~\cite{Chan2022}  } \\
         &  \OK  & \OK &     & \OK & \OK & \scriptsize{GNARF~\cite{bergman2022gnarf}} \\
     \OK &  \OK  &     & \OK   &  & &  \scriptsize{Latent3D \cite{canfes2023text},Described3D \cite{wu2023high}} \\
     \OK &  \OK  &     & \OK   &  & \OK & \scriptsize{TG-3DFace~\cite{tg_3dface_yu2023towards},DreamPortrait~\cite{cheng2023efficient}} \\
     \OK &  \OK  & \OK  & \OK   &  &   &   \scriptsize{CLIPFace \cite{aneja2023clipface}} \\
     \OK & \OK & \OK & \OK  & \OK & \OK &  \bf\scriptsize{\sysname (Ours)} \\
     \bottomrule
     \end{tabular}}
     \caption{Representative related methods in generative 3D face synthesis. Ours (\sysname) is the only method to allow text-guided synthesis of high quality 3D portraits with explicit geometry/camera control from an unlabelled 2D dataset. \vspace{-5mm} 
     }
\label{tab:relwork}
\end{table}

\vspace{-5pt}
\paragraph{3D-Aware Generative Face Synthesis.}
GANs \cite{creswell2018generative,karras2019style,karras2020analyzing} became increasingly popular in the last decade for creating high-quality photo-realistic images. Recent works have used 3D-aware multi-view consistent GANs from a collection of single-view 2D images in an unsupervised manner. The key idea is to combine differentiable rendering with 3D scene representations such as meshes, point clouds, voxels, and implicit neural representations. Among these representations, neural implicit representations \cite{Chan2022} have recently become a major focus of attention due to their superior rendering quality. Even though previous 3D-aware GANs can control camera viewpoints, they lack precise and semantically coherent control over the geometry and appearance attributes. To tackle geometry control, recent works  \cite{bergman2022gnarf,xu2023omniavatar} propose articulated generative 3D faces with 3D parametric model control. For appearance control, \cite{tg_3dface_yu2023towards} achieves text-guided 3D face generation without precise geometry control. In contrast, our method offers control over both appearance and geometry in the generation of 3D faces.


\paragraph{Text-to-3D Face Generation.}
The goal here is to produce an image that visually depicts a text description. This can be accomplished with GANs \cite{reed2016generative,zhang2017stackgan,zhang2018stackgan++,xu2018attngan,dong2017semantic,li2019controllable,zhu2019dm,tao2020df}, auto-regressive models \cite{vaswani2017attention,ramesh2021zero,ramesh2021zero,ding2021cogview,esser2021imagebart,ding2022cogview2,zhang2021m6,lee2022autoregressive,yu2022scaling} and diffusion models \cite{ho2020denoising,nichol2021glide,ramesh2022hierarchical,saharia2022photorealistic,rombach2021high}. Some works focus on text-guided facial image generation \cite{nasir2019text2facegan,stap2020conditional,xia2021tedigan,sun2021multi,wang2021faces,peng2022towards}. However, these methods only generate single-view images and do not consider 3D-aware face generation. For text-to-3D face generation, existing methods \cite{zhang2023dreamface,aneja2023clipface} build on 3D morphable face models and generate 3D faces with geometry and texture. Owing to the parametric model, these approaches can explicitly control expression, pose; however, the generation results lack shape variation. To address shape variation, TG-3DFace \cite{tg_3dface_yu2023towards} proposed text-to-face cross-modal alignment for high-quality 3D-aware face synthesis. This method has two limitations: 1) lack of explicit 3D geometry control, and 2) requirement of text annotated training dataset. Table~\ref{tab:relwork} compares existing text-to-3D face generation methods. Ours is the only method that allows text-guided synthesis of high-quality 3D portraits with explicit geometry/camera control from an unlabelled 2D dataset.

\section{Limitations and Conclusion}
\label{sec:conclusion}


\begin{figure}[t]
    \centering
    \small
    \setlength{\tabcolsep}{1pt}
    \renewcommand{\arraystretch}{0.1}
  \begin{minipage}[c]{0.3\linewidth}
  \begin{tabular}{c}
    \makecell[l]{He is a werewolf} \\
    \includegraphics[valign=m,width=\linewidth]{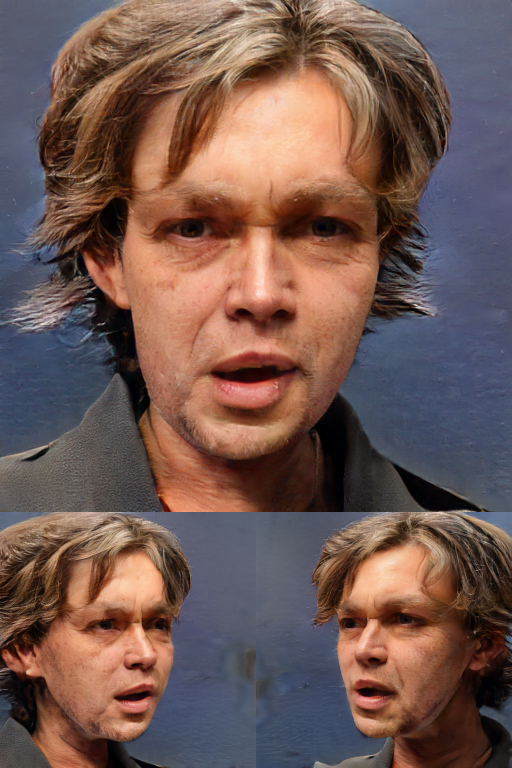} 
    \end{tabular}
    \end{minipage}
    \hspace{0.05\linewidth}
    \begin{minipage}[c]{0.49\linewidth}
    \captionof{figure}
    {
    \textbf{Limitation: Out-of-distribution Prompts.} On random FLAME parameters (producing a thin face with sideways expression).
    %
    Certain concepts such as specific artworks, celebrities, or mythical creatures that should exist in Internet-scale datasets used to train CLIP are lost, as we train our generator on a specific small subset of facial images from FFHQ.
    }
    \label{fig:limitations}
    \end{minipage}
    \vspace{-10px}
\end{figure}%

\paragraph{Limitations.} While we enable generation using prompts via an LVLM, there are bounds. As we train a generator on a specific dataset like FFHQ and we disentangle CLIP with respect to that dataset, prompting for attributes that are not represented in the data samples or identity information will not work well (Fig.~\ref{fig:limitations}). A generative approach will struggle to represent a diverse set of people if the data is not diverse, e.g., FFHQ has known skin tone biases; practical deployment must carefully avoid such biases.



\vspace{-10pt}
\paragraph{Conclusion.} Our main contribution is to identify a fundamental problem with LVLMs---more specifically with their alignment objectives---that makes them difficult to use as plug-in text-conditioners within 3D generators. Demonstrated on top of an EG3D backbone architecture, we propose regularization and canonicalisation techniques that (i) effectively disentangle 3D factors from CLIP, and (ii) avoids distribution collapse that otherwise leads to low diversity and low quality output. We demonstrate that our techniques can be important to language-guided portrait synthesis through experiments against current approaches. 





{
    \small
    \bibliographystyle{ieeenat_fullname}
    \bibliography{main}
}

\appendix
\clearpage
\section*{Appendices}
\section{Proof of the stochastic approximator of the Jacobian}
Given a vector function $f:\mathbb{R}^M\rightarrow\mathbb{R}^N$ and its variable $x\in\mathbb{R}^M$, the Jacobian $\textbf{J}\in\mathbb{R}^{N\times M}$ of $f$, we propose the following two stochastic approximators for $\left \| \textbf{J} \right \|_\text{F}^2$:
\begin{align}
    &\left \| \textbf{J} \right \|_\text{F}^2=\mathbb{E}_{v\sim\mathcal{N}(0,I_M)}\left [ \left \| \textbf{J}v \right \|^2 \right ] \label{eq:jacobiannoisevector} \\
    &\left \| \textbf{J} \right \|_\text{F}^2=\lim_{\sigma\rightarrow0}\mathbb{E}_{\epsilon\sim\mathcal{N}(0,\sigma^2I_M)}\left [ \frac{1}{\sigma^2}\left \| f(x+\epsilon)-f(x) \right \|^2 \right ] \label{eq:jacobianfinitedifference}
\end{align}
Our proofs follow the same principles as the numerical approximation of the Hessian matrix norm in~\cite{conae}. While evaluating $\left \| \textbf{J} \right \|_\text{F}^2$ naively using torch.autograd.functional.jacobian is impractically slow, \Cref{eq:jacobiannoisevector} can be efficiently evaluated by torch.autograd.functional.jvp. \Cref{eq:jacobianfinitedifference} is an even faster finite difference approximation that we use in the main paper for $R_{\textbf{r}_{\hat{x}}}$.


\paragraph{Proof for \Cref{eq:jacobiannoisevector}.} Note that for $i\neq j$, $\mathbb{E}\left [ v_iv_j \right ]=\mathbb{E}\left[v_i\right]\mathbb{E}\left[v_j\right]=0$. For $i=j$, $\mathbb{E}\left [ v_iv_j \right ]=\mathbb{E}\left [ v_i^2 \right ]=1$.
\begin{align}
&\mathbb{E}_{v\sim\mathcal{N}(0,I_M)}\left [ \left \| \textbf{J}v \right \|^2 \right ]\\
&=\mathbb{E}\left [ \left \langle \sum_{i=1}^{M}v_i\frac{\partial f}{\partial x_i},\sum_{i=1}^{M}v_i\frac{\partial f}{\partial x_i} \right \rangle \right ] \\
&=\sum_{i=1}^{M}\sum_{j=1}^{M}\mathbb{E}\left [ v_iv_j \right ]\left \langle \frac{\partial f}{\partial x_i},\frac{\partial f}{\partial x_j} \right \rangle \\
&=\sum_{i=1}^{M}\left \langle \frac{\partial f}{\partial x_i},\frac{\partial f}{\partial x_i} \right \rangle\\
&=\left \| \textbf{J} \right \|_\text{F}^2
\end{align}

\paragraph{Proof for \Cref{eq:jacobianfinitedifference}.} Apply Taylor expansion to $f(x+\epsilon)$ at $x$:
\begin{equation}
    f(x+\epsilon)=f(x)+\textbf{J}\epsilon+R(x,\epsilon)
\end{equation}
where
\begin{equation}
    R(x,\epsilon)=\sum_{K=2}^{\infty}\frac{1}{K!}\sum_{i_1,\dots,i_K}\epsilon_{i_1}\dots\epsilon_{i_K}\frac{\partial^Kf}{\partial x_{i_1},\dots,\partial x_{i_K}}
\end{equation}
which gives us
\begin{align}
    &\lim_{\sigma\rightarrow0}\mathbb{E}_{\epsilon\sim\mathcal{N}(0,\sigma^2I_M)}\left [ \frac{1}{\sigma^2}\left \| f(x+\epsilon)-f(x) \right \|^2 \right ]\\
&=\lim_{\sigma\rightarrow0}\mathbb{E}\left [ \frac{1}{\sigma^2}\left \| \textbf{J}\epsilon+R(x,\epsilon) \right \|^2 \right ]\\
\nonumber&=\lim_{\sigma\rightarrow0}\mathbb{E}\left [ \frac{1}{\sigma^2}\left \| \textbf{J}\epsilon \right \|^2 \right ]+\lim_{\sigma\rightarrow0}\mathbb{E}\left [ \frac{2}{\sigma^2} \left\langle \textbf{J}\epsilon,R(x,\epsilon) \right\rangle \right ]+\\&\,\,\,\,\,\,\,\lim_{\sigma\rightarrow0}\mathbb{E}\left [ \frac{1}{\sigma^2} \left\langle R(x,\epsilon),R(x,\epsilon) \right\rangle \right ]\\
\nonumber&=\left \| \textbf{J} \right \|_\text{F}^2+\lim_{\sigma\rightarrow0}\mathbb{E}\left [ \frac{2}{\sigma^2} \left\langle \textbf{J}\epsilon,R(x,\epsilon) \right\rangle \right ]+\\&\,\,\,\,\,\,\,\lim_{\sigma\rightarrow0}\mathbb{E}\left [ \frac{1}{\sigma^2} \left\langle R(x,\epsilon),R(x,\epsilon) \right\rangle \right ]\ .
\end{align}
We now analyze $\epsilon_{i_1}\dots\epsilon_{i_K}$, which can be rewritten to $\epsilon_1^{p_1}\dots\epsilon_M^{p_M}$ where $p_1+\cdots+p_M=K$. Note that $\left\langle \textbf{J}\epsilon,R(x,\epsilon) \right\rangle$ and $\left\langle R(x,\epsilon),R(x,\epsilon) \right\rangle$ imply $K\geqslant3$ as we have $K\geqslant2$ from $R(x,\epsilon)$. In such scenario,
\begin{itemize}
    \item Either there is at least one odd element in $\{p_1,\dots,p_M\}$ and $\mathbb{E}\left[ \epsilon_{i_1}\dots\epsilon_{i_K} \right]=0$ because odd moments of a zero mean Gaussian are zero, 
    \item Or $p_1,\dots,p_M$ are all even but $\mathbb{E}\left[ \epsilon_{i_1}\dots\epsilon_{i_K} \right]$ is a higher order infinitesimal than $\sigma^2$, and we have $\lim_{\sigma \to 0} \frac{1}{\sigma^2}\mathbb{E}\left[ \epsilon_{i_1}\dots\epsilon_{i_K} \right]=0$.
\end{itemize}   
Therefore:
\begin{align}
    &\lim_{\sigma\rightarrow0}\mathbb{E}\left [ \frac{2}{\sigma^2} \left\langle \textbf{J}\epsilon,R(x,\epsilon) \right\rangle \right ]\\&=\lim_{\sigma\rightarrow0}\mathbb{E}\left [ \frac{1}{\sigma^2} \left\langle R(x,\epsilon),R(x,\epsilon) \right\rangle \right ]=0. \hspace{10pt} \blacksquare
\end{align}
To implement $R_{\textbf{r}_{\hat{x}}}$, we empirically set $\sigma=0.1$. We experimented with using 1, 2, 4 and 8 $\epsilon$ vectors for the expectation, but found no significant difference and thus settled for using only one $\epsilon$ vector to minimize the computation cost.

\section{Implementation Details}
\label{sec:implementation_details}

Our implementation extends the official PyTorch implementation\footnote{https://github.com/NVlabs/eg3d} of EG3D \cite{cheng2023efficient}. We initialize the generator backbone with the weights from the official checkpoint provided by the authors. For the most part, we follow the training strategies of EG3D. However, we drop the second training stage of EG3D and fix neural rendering resolution to 64$\times$64 due to computation budget limitations. We opt for lower initial learning rates $\gamma=0.001$ for both generator $G$ and discriminator $D$ rather than 0.0025 for $G$ and 0.002 for $D$. We also remove the generator pose conditioning, as mentioned in the main paper.

For deformation, we use the surface-field implementation from GNARF by Bergman et al.~
\cite{bergman2022gnarf}. For faster deformation computation, we simplify the FLAME mesh to 2500 triangles. However, we use the full mesh for generating the mesh render $rdr$ in discriminator conditioning.

For objectives, we follow EG3D and use a non-saturating adversarial loss~\cite{gan}, $R_1$ gradient penalty~\cite{r1}, and density regularization~\cite{cheng2023efficient}. We additionally append our regularizations $R_{\textbf{r}_{\hat{x}}}$ and $R_\text{norm}$ to the training objectives, and empirically set their weights to 0.01 and 10 respectively.

\begin{figure*}
\scalebox{0.90}{
\small
    \setlength{\tabcolsep}{1pt}
    \renewcommand{\arraystretch}{0.4}
  \begin{tabular}{cccccc}
    Input & Red Eyes & Blue Eyes & Long Hair & Blue Hair & Blonde \\ 
    
    \includegraphics[valign=m,width=0.18\textwidth]{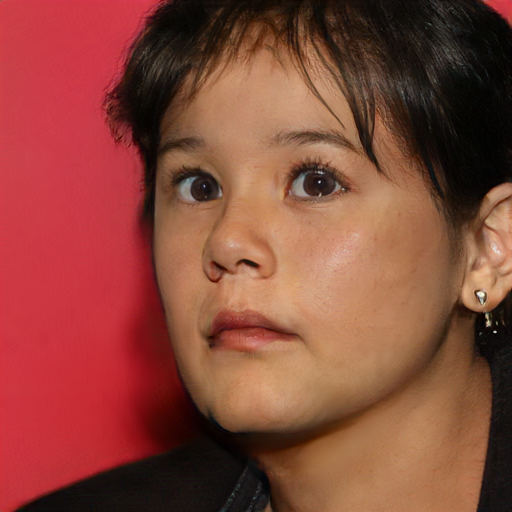} & 
    \includegraphics[valign=m,width=0.18\textwidth]{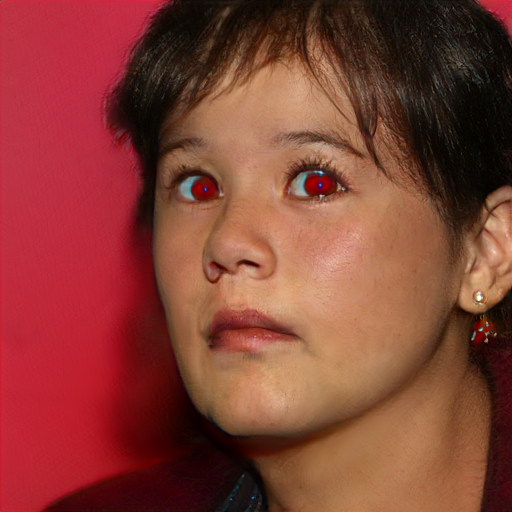} &  
    \includegraphics[valign=m,width=0.18\textwidth]{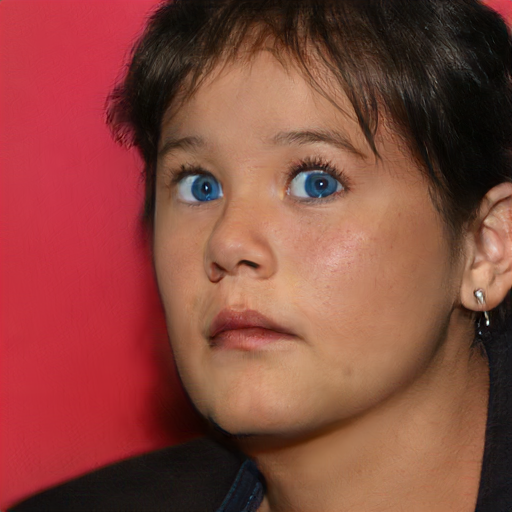} &
    \includegraphics[valign=m,width=0.18\textwidth]{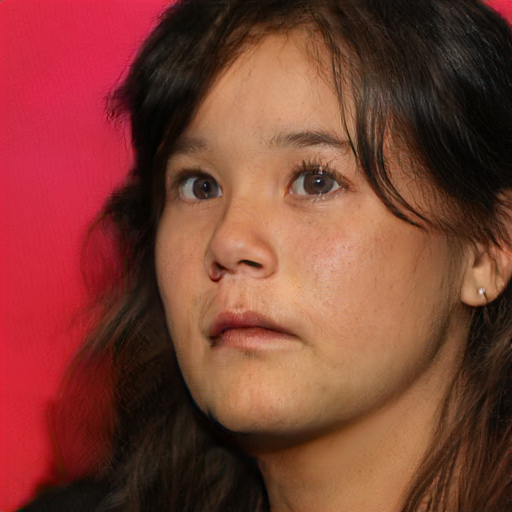} &
    \includegraphics[valign=m,width=0.18\textwidth]{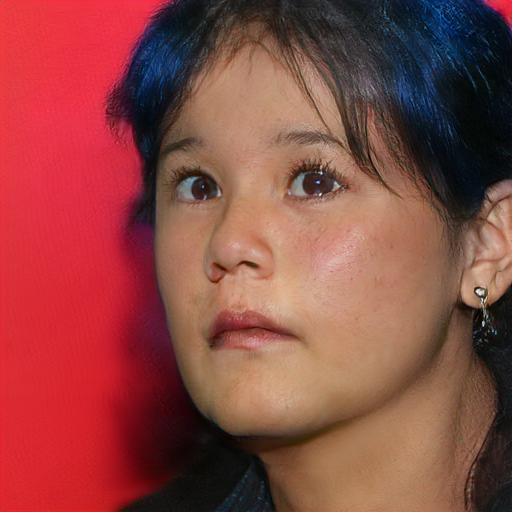} &
    \includegraphics[valign=m,width=0.18\textwidth]{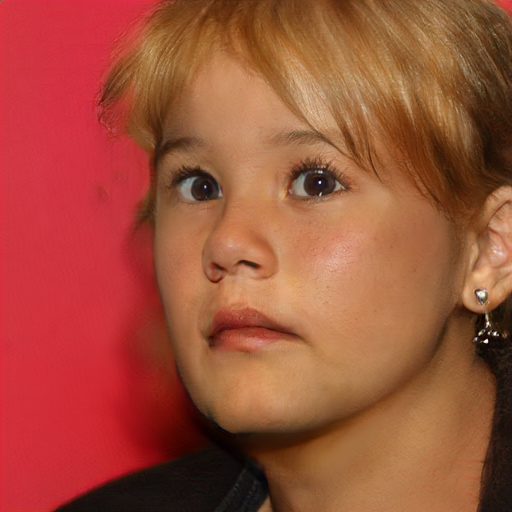} \\
    
    \includegraphics[valign=m,width=0.18\textwidth]{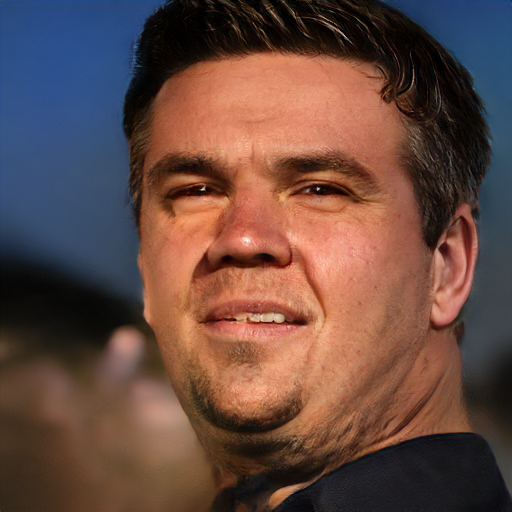} & 
    \includegraphics[valign=m,width=0.18\textwidth]{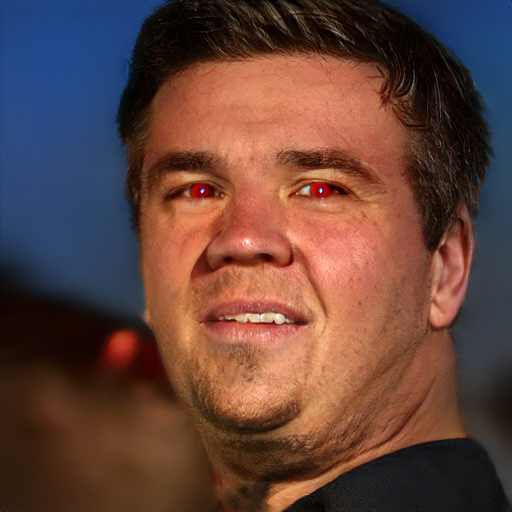} &  
    \includegraphics[valign=m,width=0.18\textwidth]{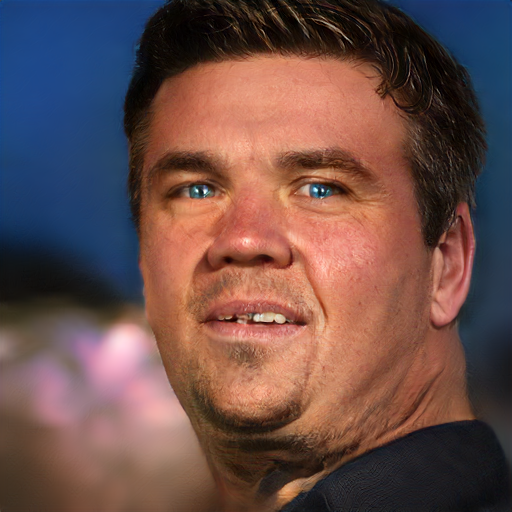} &
    \includegraphics[valign=m,width=0.18\textwidth]{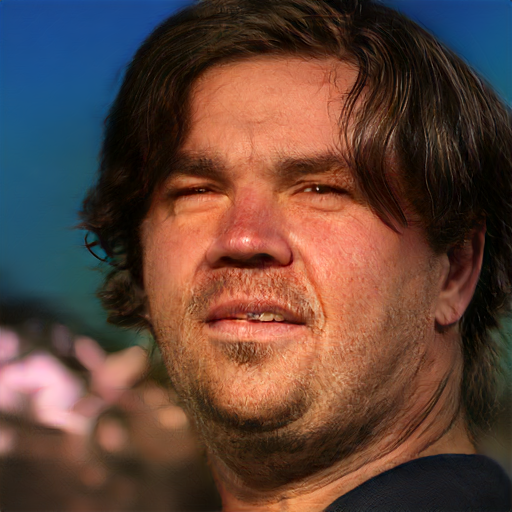} &
    \includegraphics[valign=m,width=0.18\textwidth]{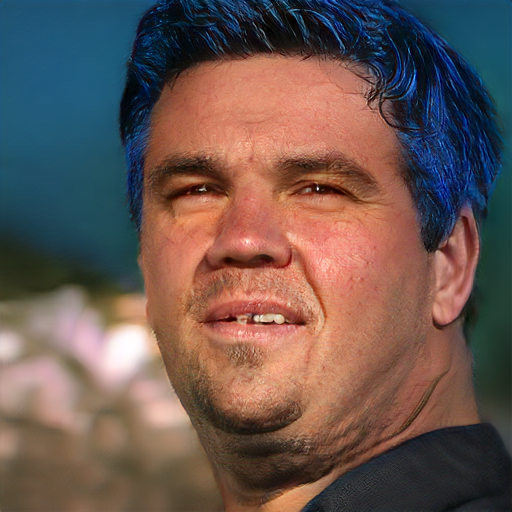} &
    \includegraphics[valign=m,width=0.18\textwidth]{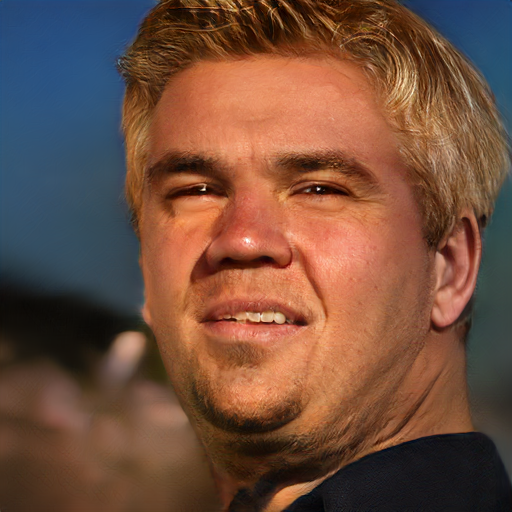} 
    

    
    
  \end{tabular}
}
    \caption{Text-guided appearance editing results.}
    \label{fig:application_edit}
\end{figure*}

\section{Text-Guided Appearance Editing.}
In addition to text conditional generation, our model is also capable of fine-grained editing of existing portraits. The nature of editing is no different from alignment: we can adopt a specialized alignment network $E_t$ to align any style vector $w$ to our editing request $t$ similarly to Equation (11) in the main paper:
\begin{align}
    w_t(\alpha)=w+\alpha E_t(w)
\end{align}
$t$ is represented by text prompts such as ``blonde'', ``blue eyes'', etc., and $\alpha$ controls the editing strength. We hypothesize that the $\mathcal{W}^+$ space is sufficiently disentangled for local editing, i.e., it is achievable to apply the desired effect while preserving the rest of the portrait by manipulating $w$. A similar design has been proposed by Aneja et al.~\cite{aneja2023clipface} in the context of textured 3DMM editing.

The training scheme of $E_t$ remains a question. While it is possible to optimize $E_t$ using an adversarial objective as with $T_G$, the problem with the adversarial loss is that it encourages alignment but not disentanglement. There is no incentive to preserve the content of the initial portrait while applying the desired effect, thus making it unsuitable for editing. To promote disentanglement, we adopt the contrastive CLIP loss from Aneja et al.~\cite{aneja2023clipface} that allows us to single out the components related to $t$ and modify them. We render the initial portrait and the edited portrait under a neutral camera pose and 3DMM parameters. Then, the change induced by $E_t$ in the CLIP embedding space is given by:
\begin{small}
\begin{align}
    \Delta x=E_\text{img}(G(w_t,\textbf{c}_\text{n\_cam},\textbf{c}_\text{n\_geo}))-E_\text{img}(G(w,\textbf{c}_\text{n\_cam},\textbf{c}_\text{n\_geo}))
\end{align}
\end{small}
Meanwhile, we define a template prompt $\bar{t}=\text{``a photo of a neutral face''}$ and the expected change induced by $t$ is:
\begin{align}
    \Delta t=E_\text{txt}(t)-E_\text{txt}(\bar{t})
\end{align}
We freeze $G$ and train $E_t$ with the objective of aligning $\Delta x$ to $\Delta t$. We additionally apply $R_\text{norm}$: without norm growth regularization, we observe instant collapse early in the training. The full training objective is:
\begin{align}
    1-\frac{\Delta x\Delta t}{\left \| \Delta x \right \|\left \| \Delta t \right \|}+\eta R_\text{norm}
\end{align}
where $\eta$ is empirically set to 10. We visualize our editing results in Figure \ref{fig:application_edit}. 
Although a similar training objective is adopted at inference time in TG-3DFace~\cite{tg_3dface_yu2023towards}, our editing results are considerably more realistic and disentangled as we do not update $G$ (main paper Figure 9).

\end{document}